\documentclass[10pt,twocolumn,letterpaper]{article}

\usepackage{cvpr}
\usepackage{times}
\usepackage{epsfig}
\usepackage{graphicx}
\usepackage{amsmath}
\usepackage{amssymb}
\usepackage{booktabs}
\usepackage{hyperref}
\usepackage{caption}
\usepackage{wrapfig}
\usepackage{array}
\usepackage{geometry} 
\usepackage{multicol}
\usepackage{multirow}
\cvprfinalcopy 

\def\cvprPaperID{****} 

\begin{document}

\title{Towards Optimal Convolutional Transfer Learning Architectures for Breast Lesion Classification and ACL Tear Detection}

\author{Daniel Frees\\
Stanford University\\
Department of Statistics\\
{\tt\small dfrees@stanford.edu}
\and
Moritz Bolling\\
Stanford University\\
Department of Statistics\\
{\tt\small mbolling@stanford.edu}
\and
Aditri Bhagirath\\
Stanford University\\
Department of Computer Science\\
{\tt\small aditri@stanford.edu}
}

\maketitle

\begin{abstract}
   Modern computer vision models have proven to be highly useful for medical imaging classification and segmentation tasks, but the scarcity of medical imaging data often limits the efficacy of models trained from scratch. Transfer learning has emerged as a pivotal solution to this, enabling the fine-tuning of high-performance models on small data. Mei et al. (2022) \cite{Mei2022RadImageNet} found that pre-training CNNs on a large dataset of radiologist-labeled images (RadImageNet) enhanced model performance on downstream tasks compared to ImageNet pretraining. The present work extends Mei et al. (2022) by conducting a comprehensive investigation to determine optimal CNN architectures for breast lesion malignancy detection and ACL tear detection, as well as performing statistical analysis to compare the effect of RadImageNet and ImageNet pre-training on downstream model performance. Our findings suggest that 1-dimensional convolutional classifiers with skip connections, ResNet50 pre-trained backbones, and partial backbone unfreezing yields optimal downstream medical classification performance. Our best models achieve AUCs of 0.9969 for ACL tear detection and 0.9641 for breast nodule malignancy detection, competitive with the results reported by Mei et al. (2022) and surpassing other previous works \cite{sun2020deep}. We do not find evidence confirming RadImageNet pre-training to provide superior downstream performance for ACL tear and breast lesion classification tasks.
\end{abstract}

\section{Introduction}
\label{sec:intro}

Computer vision models have the power to enable enhanced detection accuracy, better patient triage, and high-quality automated segmentation for medical imaging (\cite{bressem2020comparing}, \cite{Mei2022RadImageNet}, \cite{sun2020deep}, \cite{Rose1996}, \cite{Kim2022}), but data collection challenges often necessitate model development with small datasets. Traditionally, small data would provide insufficient learning power to train large convolutional neural networks (CNNs), but a recent paradigm shift to transfer learning mitigates this: by pre-training computer vision models on massive datasets and then using these pre-trained weights as feature extractor inputs into a smaller classifier layer, fine-tuning can be performed on small datasets, yielding local minima with better generalization performance \cite{Zhao2023}. In the field of computer vision, transfer learning typically begins with pre-training feature extractor weights using the massive ImageNet classification challenge dataset.

Mei et al. (2022) \cite{Mei2022RadImageNet} propose that pre-training on 1.35 million radiologist-labeled radiological images might improve fine-tuning performance for downstream medical imaging tasks, considering radiology's distinct objects and artifacts compared to the natural images ImageNet \cite{deng2009imagenet} was trained on, finding that RadImageNet weights improve average downstream classification and segmentation performance. However, no details were provided as to the exact neural network architectures used to conduct these statistical analyses. For a full analysis of the original work, see \autoref{sec:critique}.



We seek to extend the work of Mei et al. (2022) \cite{Mei2022RadImageNet} by elucidating the best architectural choices for developing transfer-learned CNN models for two radiology tasks: anterior cruciate ligament (ACL) tear detection \autoref{tab:acl_tear_detection} and breast nodule malignancy detection \autoref{tab:breast_cancer_detection}. 

\begin{table}[!h]
    \small
    \centering
    \begin{tabular}{@{}p{1.5cm}p{6cm}@{}} 
        \toprule
        \textbf{Task} & \textbf{Details} \\
        \midrule
        \textbf{Dataset} & MRNet dataset \cite{StanfordML2023MRNet} \\
        \textbf{Description} & Includes 1021 ACL knee MRI exams performed at Stanford University Medical Center. Labels extracted manually from clinical reports. 570 abnormal (ACL tear) exams. \\ 
        \textbf{Input} & \texttt{.png} MRI image \\
        \textbf{Output} & label (\texttt{y/n}) for meniscus tear \\
        \bottomrule
    \end{tabular}
    \caption{Details of the ACL tear detection task}
    \label{tab:acl_tear_detection}
\end{table}

\begin{table}[!h]
    \small
    \centering
    \begin{tabular}{@{}p{1.5cm}p{6cm}@{}} 
        \toprule
        \textbf{Task} & \textbf{Details} \\
        \midrule
        \textbf{Dataset} & Kaggle Breast Ultrasound Image dataset \cite{Shah2021BreastUltrasound} \\
        \textbf{Description} & Includes 780 \texttt{.png} images from 600 female patients. \\
        \textbf{Input} & \texttt{.png} US image \\
        \textbf{Output} & label (\texttt{benign/malignant/normal}) \\
        \bottomrule
    \end{tabular}
    \caption{Details of the malignant breast cancer detection task}
    \label{tab:breast_cancer_detection}
\end{table}

We experiment with various classifier architectures, unfreezing strategies, optimizers, and hyperparameters and evaluate their effect on classification performance. We also use gradient-weighted Class Activation Mapping (Grad-CAM)  \cite{selvaraju2019grad} to qualitatively interpret the diagnostic decisions made by our best models. Lastly, we aim to validate the original results suggesting that RadImageNet pretraining yields more robust generalization performance for radiology tasks. 

Overall, we find that convolutional classifier layers with skip connections, $5$-layers of backbone unfreezing, SGD with momentum and strong weight decay, and ImageNet backbone initialization yields superior detection models, with our best breast model achieving a test set area under the receiver operating characteristic curve (AUC) of 0.9641 and our best ACL modeling achieving a test AUC of 0.9969. 


\section{Related Work}

\subsection{Model Efficacy and State of the Art}

Previous works have demonstrated that transfer learning is highly effective for developing convolutional medical imaging models on small datasets (\cite{Mei2022RadImageNet}, \cite{bressem2020comparing}, \cite{Kim2022}). The present work seeks to extend Mei et al. (2022) \cite{Mei2022RadImageNet} by more deeply understanding the best architecture choices for medical diagnostics transfer learning, specifically for diagnosing ACL tears and breast lesion malignancy. 

ACL tear detection is a fairly easy task for clinicians, where the main challenge is locating the ACL and determining whether it is properly attached, or invisible on the MRI. Clinicians achieve 99\% detection accuracy for full ACL tears \cite{Rose1996}. Breast lesion malignancy detection is harder, with a 2021 study finding breast lesion malignancy detection AUCs of 0.855 for their machine learning models, and 0.805 average AUC for the radiologists included in the study \cite{sun2020deep}.

\subsection{Medical Classification Architectures}

Other works have previously investigated architecture best practices for medical imaging. Kim et al. (2022) \cite{Kim2022} perform a literature review of transfer learning for medical imaging applications, finding that ResNet50 and InceptionV3, limited backbone layer unfreezing, and small learning rates yield the best performance. Bressem et al. (2020) \cite{bressem2020comparing} find that many transfer-learned ResNet50 and DenseNet121 models achieve high performance on both chest radiograph classification and COVID-19 detection. In their 2024 study, Mathivanan et al. (2024) \cite{mathivanan2024employing} explore the efficacy of deep transfer learning for brain tumor diagnosis using MRI, employing architectures like ResNet152, VGG19, DenseNet169, and MobileNetv3 to predict four tumor categories. They achieve a staggering 99.75\% accuracy and highlight the importance of image enhancement and partial backbone unfreezing, both of which we employ in the present work. 




\subsection{Statistical Evaluation}

One of our goals is to statistically evaluate performance differences between RadImageNet and ImageNet pre-training, towards validating the original RadImageNet work, which found that RadImageNet weights were superior to ImageNet weights for downstream medical classification and segmentation tasks \cite{Mei2022RadImageNet}. Mei et al. (2022) showed that in thyroid nodule, breast mass classification and anterior cruciate ligament (ACL) injury detection tasks, RadImageNet models achieved improvements in AUC by up to 9.4\%. Lesion localization accuracy, as measured by Dice scores, was notably enhanced, with gains of up to 64.6\% in thyroid nodule detection. Towards comparing the performance of our RadImageNet and ImageNet models, we employ the nonparametric DeLong test \cite{DeLong1988} to compare fine-tuned AUCs, and use linear mixed effect models \cite{Bates2015} to robustly estimate effects given that our observations are not independent.


\subsection{Interpretability} 

Especially in the medical field, model interpretability can be just as important as performance metrics. Late convolutional layers in computer vision models have been shown to extract high-level spatial features \cite{zhang2020} from input images. Grad-CAM, introduced by Selvaraju et. al. (2019) \cite{selvaraju2019grad}, uses this fact to enable visualization of important pixel region heatmaps in input images, measuring their effect on model decisions via class activation gradient mappings. Panwar (2020) \cite{Panwar2020} shows that these Grad-CAM visualizations can be useful for medical imaging model interpretation for pneumonia and COVID-19 detection.


\section{Methods}
\label{sec:methods}

Our model architectures and training frameworks were constructed using PyTorch \cite{paszke2019pytorch}. All experiments were run on an Apple Silicon M3 Max GPU with 30 cores and 36 GB unified memory (see \autoref{sec:hardware} for details).

\subsection{Transfer Learning Approach}


Transfer learning can be formalized as follows. A \textbf{domain} $D$ is characterized by a feature space $\mathcal{X}$ and a marginal probability distribution $P(X)$, where $X \subseteq \mathcal{X}$. A \textbf{task} $T$ associated with a domain involves a label space $\mathcal{Y}$ and an objective predictive function $f(\cdot)$. Given a source domain $D_S$ with a learning task $T_S$, and a target domain $D_T$ with a learning task $T_T$, transfer learning aims to enhance the learning of the predictive function $f_T(\cdot)$ in $D_T$ by leveraging the knowledge from $D_S$ and $T_S$.

To establish our source domain knowledge ($D_S$, $T_S$) we downloaded both pre-trained RadImageNet weights and pre-trained ImageNet weights (v1) for InceptionV3, ResNet50, and DenseNet121. Each set of weights was pre-trained on large-scale multi-class classification tasks, for radiological images and real-world images respectively. ImageNet spans 1000 object classes from a diverse domain, and contains about 1.3 million training images, 50,000 validation images and 100,000 test images \cite{imagenet_dataset}. The RadImageNet dataset is much more specific, consisting of 1.35 million annotated medical images including CT scans, MRI and ultrasound data. 

\begin{figure}[h]
    \centering
    \includegraphics[width=0.2\textwidth]{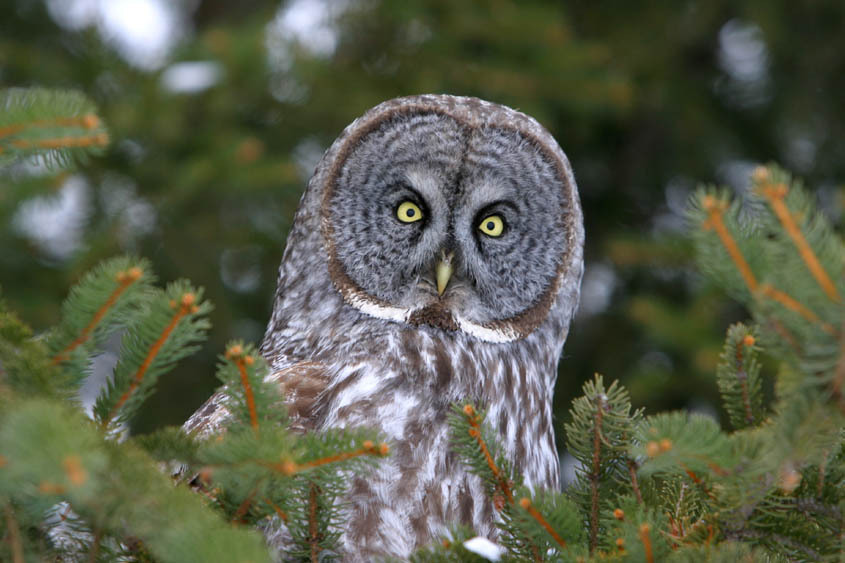}
    \includegraphics[width=0.2\textwidth]{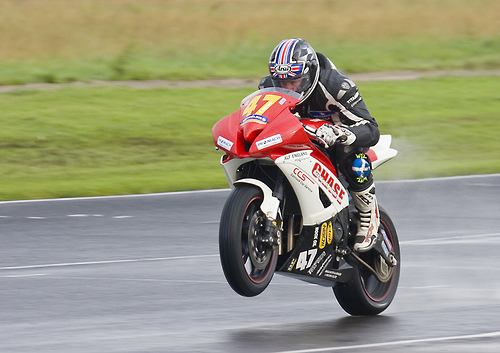}
    \caption{Two examples of images from ImageNet}
    \label{fig:owl_helmet_horizontal}
\end{figure}

\begin{figure}[!h]
    \centering
    \includegraphics[width=0.2\textwidth]{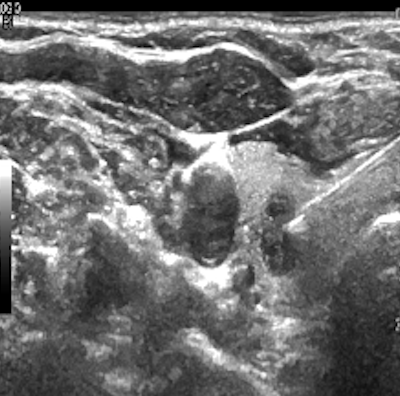}
    \includegraphics[width=0.2\textwidth]{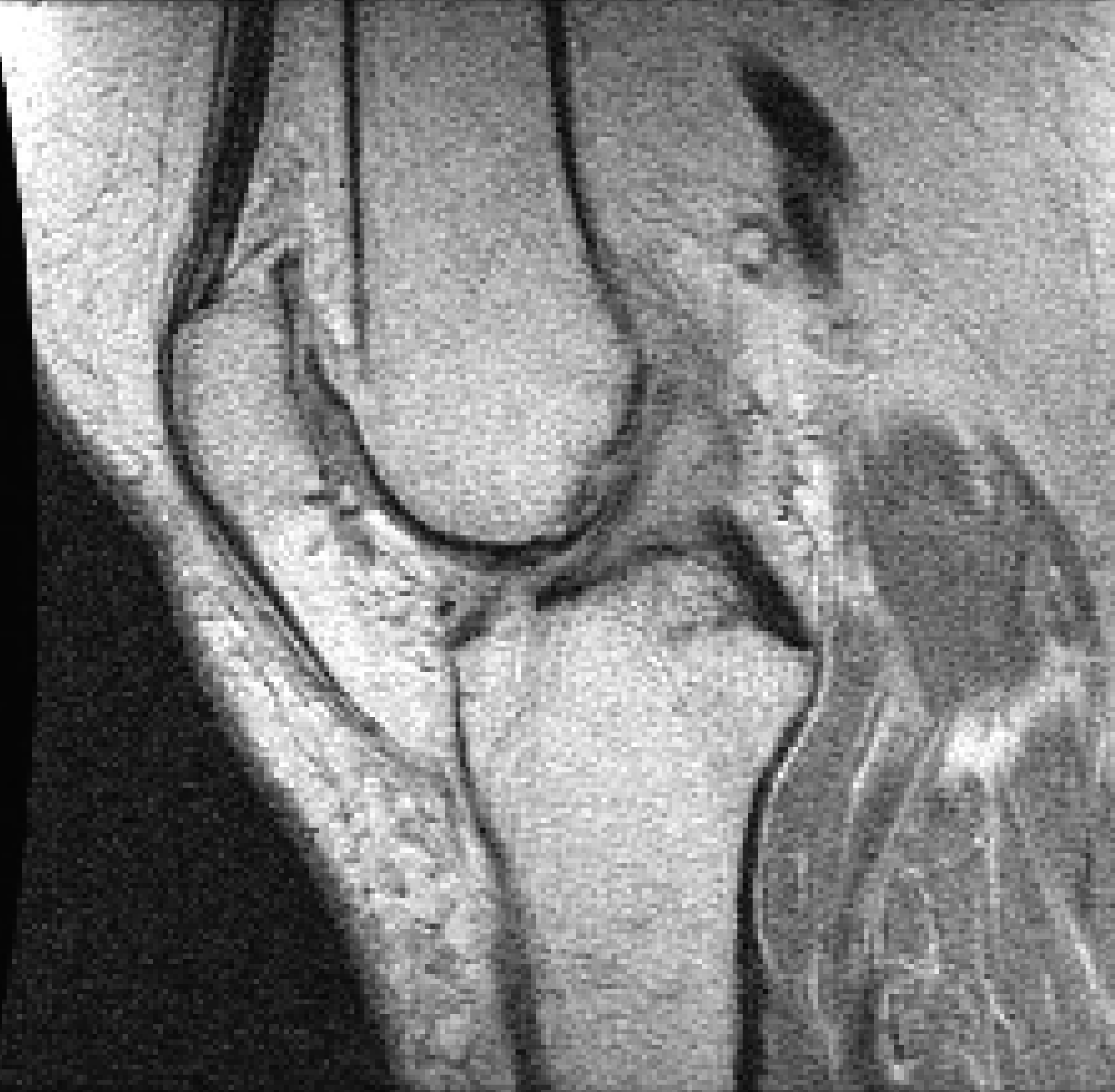}
    \caption{Two examples of images from RadImageNet Left: Thyroid nodule ultrasound, Right: Knee ACL MRI. Note that RadImageNet does not contain any breast pathology images.}
    \label{fig:radimage_examples}
\end{figure}

See \autoref{fig:owl_helmet_horizontal} for image examples from ImageNet and \autoref{fig:radimage_examples} for examples from RadImageNet

To use pre-trained RadImageNet models, we designed a 'Backbone' PyTorch module to accomodate three large convolution image models (InceptionV3, ResNet50, DenseNet121), with the final layer removed to expose feature weights rather than predictions. We then concatenated this Backbone with a variety of Classifier layers, rendering the Backbone a feature extractor. For a full visualization of our model architecture, see \autoref{sec:transfer_architecture}.

All of our Classifier layer architectures are described below: 

\begin{enumerate}
 
    \item \textbf{Linear:} A single fully connected linear layer mapping backbone features to 2 output classes.
    \item \textbf{Non-Linear:} Comprises two linear layers, a leaky ReLU activation, dropout for regularization, and batch normalization after the first layer.
    \item \textbf{Convolution:} Includes a 1D convolutional layer and two fully connected layers, with batch normalization and dropout applied after the convolution and the first linear layer. Leaky ReLU activation ensures effective gradient flow.
    \item \textbf{Convolution with Skip Connections:} Builds on the convolution model by adding a skip connection that merges Conv and skip pathway outputs via element-wise addition, enhancing gradient flow and preserving information through deeper layers. This model is expected to be the most effective, combining the convolutional model's expressivity with simpler gradient propagation.

\end{enumerate}

We validated our initial PyTorch implementation by refining the TensorFlow code and conducting baseline experiments with Linear classifiers on InceptionV3 models (detailed in \autoref{sec:baselines}). 

Our models employed Kaimeng initialization to prevent gradient issues, as described in \cite{He2015}. We optimized our training using a variety of optimizers (Adam, AdamW, RAdam, and SGD with momentum) and weight decay strengths, applying learning rate decay techniques such as cosine annealing and beta decay after observing noisy validation performance. Cross-entropy loss was chosen for classification tasks to enhance accuracy and AUC, with decision thresholds adaptable based on clinical needs.

Unlike the original RadImageNet framework, we introduced the ability to unfreeze variable numbers of layer groups, adapting the training process to be more flexible and potentially more effective, as evidenced by recent studies like Mathivanan et al. (2024) \cite{mathivanan2024employing} which demonstrated the efficacy of partial unfreezing in MRI brain tumor diagnosis. This approach allowed us to unfreeze the last five layer groups of a ResNet50 backbone, enhancing model adaptability and performance. Note that these layer groups can be, for example, \texttt{torch.nn.Sequential}s of multiple ResNet50 \texttt{'Bottleneck'} layers each consisting of several Convs, Batchnorms, and MaxPools.


In summary, our transfer learning optimization problem for each target domain $D_T$ and task $T$ can be stated as:

\[
\min_{\theta} \mathcal{L}(\theta) = -\frac{1}{N} \sum_{i=1}^{N} \left[ y_i \log(p_i) + (1 - y_i) \log(1 - p_i) \right]
\]

where

\[
p_i = \frac{\exp(f_\theta(x_i))}{\sum_{c=0}^{1} \exp(f_\theta(x_i)_c)}
\]

and \( f_\theta(x_i)_c \) is the output of our model for class \( c \) with parameters \(\theta\) for input \(x_i\), and \(y_i\) is the ground truth label ($0$ or $1$). $c=1$ corresponds to positive diagnoses (\texttt{'malignant'}/\texttt{'yes'}). $f_\theta$ represents the entire combined Backbone + Classifier model, noting that the number of learnable parameters in  $\theta$ depends on the amount of backbone unfreezing enabled. 

\subsection{Evaluation}

Following (Mei et al. 2022) \cite{Mei2022RadImageNet}, we utilized AUC as the primary metric for comparing experimental outcomes. Consequently, for statistical analyses comparing models we employed the DeLong test, a nonparametric method that allows for the comparison of correlated ROC curves without presupposing the distribution of the underlying data \cite{DeLong1988}. The analysis was completed in R \cite{R2024} using the ``pROC" package \cite{pROC2011}. We further compared the predictions of models using McNemar tests \cite{mcnemar1947} to determine whether the marginal probabilities of two classifiers are equal. To robustly evaluate the impact of pretraining and classifier architecture on the AUC, in the face of non-independent observations, we used mixed effect models with AUC as the dependent variable, parameter of interest as the fixed effect, and other parameters as the random effects. While the linear modeling assumptions inherent to these tests are unlikely to fully capture the relationships in question, these tests nonetheless provide valuable insights into the directionality and magnitude of hyperparameter effects. This analysis was conducted using the ``lme4" package \cite{Bates2015}. Lastly, we used Wilcoxon rank sum tests \cite{Wilcoxon1945} to compare distributions of our data. Given the dependence structure of our observations, this test will be biased toward rejecting the null hypothesis and must be interpreted with care. A non-rejection can still be a valuable insight.

\subsection{Visual Explainability}

To enable visual explainability for our diagnostic model decisions we utilized Grad-CAM to visualize input pixel importances as measured by gradient-based positive influence on the ground truth label score for each image. Selvaraju et al. (2019) \cite{selvaraju2019grad} propose a framework called Gradient-weighted Class Activation Mapping (Grad-CAM) which enables visual interpretation of convolutional neural networks (CNNs) We first computed the gradient of the score for target class $c$, $y^c$ (before the softmax), with respect to feature map activations $A^k$ of previous convolutional layers, i.e., $\frac{\partial y^c}{\partial A^k_{ij}}$. While flowing gradients back to the 'target' convolutional layer, upstream gradients were global-average-pooled over the spatial dimensions. A weighted combination of forward activation maps was conducted at the target layer to obtain a coarse heatmap of spatial feature importances in the same size as the convolutional feature maps at the target layer, and ReLU was applied to only retain information about features which \textit{increased} the probability of target class $c$. We adopted the original grad-CAM method by selecting the second to last and third to last ResNet50 layer groups (each consisting of numerous Conv, BatchNorm, Maxpool layers) as our target layers and averaging Grad-CAM results across these layers (our code also supports DenseNet121 interpretation, using the entirety of 'DenseBlock4' as our target layer). This falls in line with the suggestions of Selvaraju et al. (2019) \cite{selvaraju2019grad} who propose and validate empirically that late convolutional layers best capture high-level spatial information related to class activations. Our target class was always the predicted label for the image. 

\section{Data}


\subsection{Fine-tuning data}

\begin{figure}[!h]
    \centering
    \includegraphics[width=0.2\textwidth]{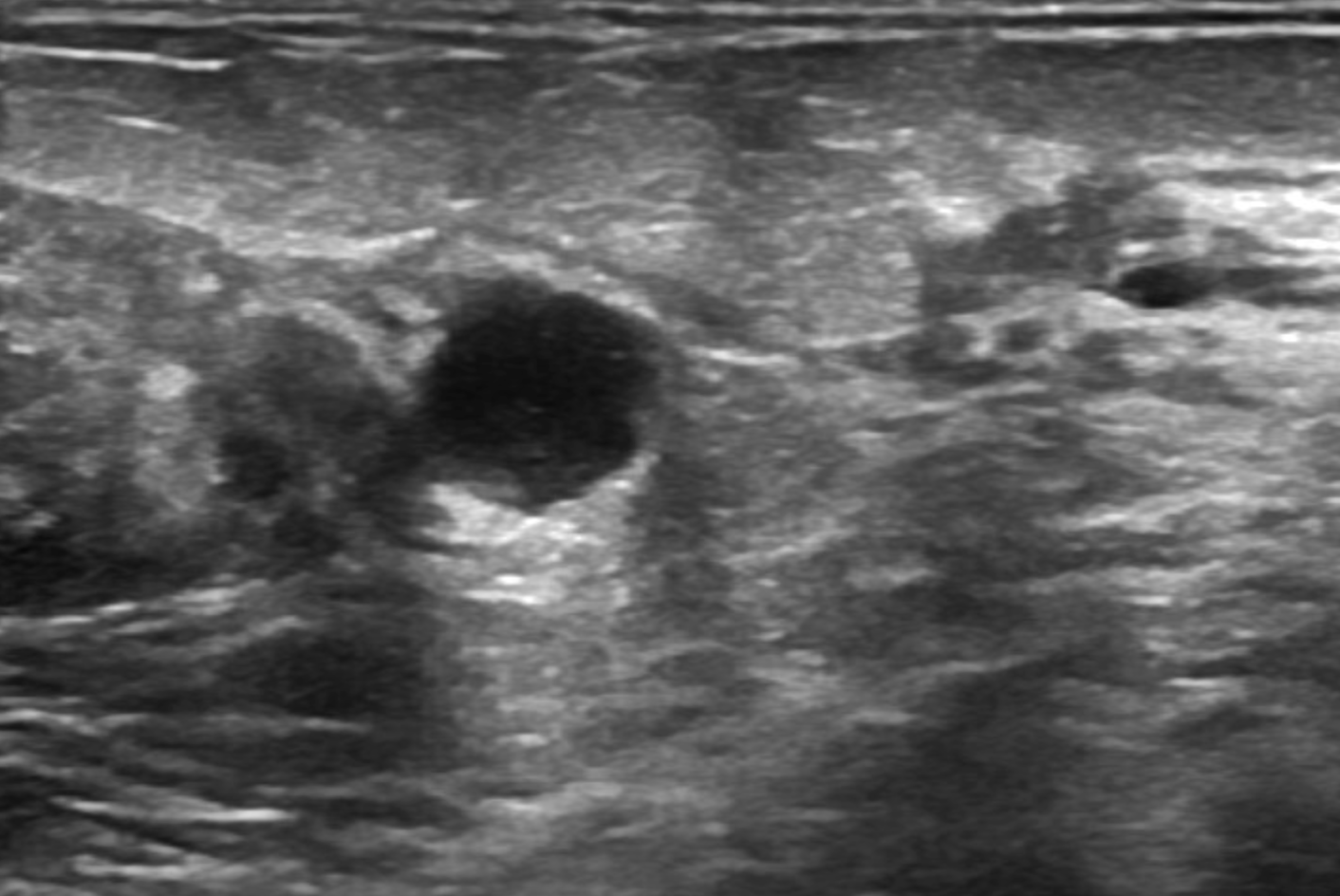}
    \includegraphics[width=0.2\textwidth]{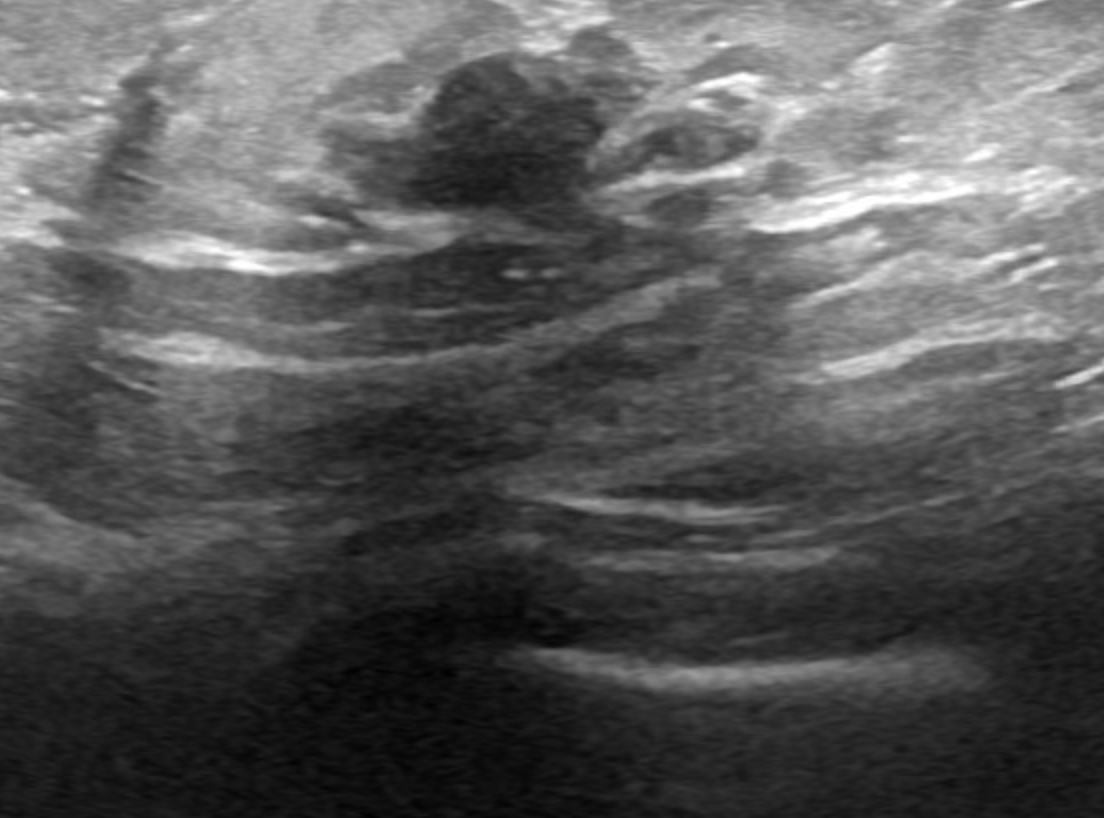}
    \caption{Instances of benign (left) and malignant (right) breast nodule ultrasounds from the Kaggle fine-tuning dataset for our downstream breast cancer nodule classification task.}
    \label{fig:breast_ultrasound_examples}
\end{figure}

\begin{figure}[!h]
    \centering
    \includegraphics[width=0.2\textwidth]{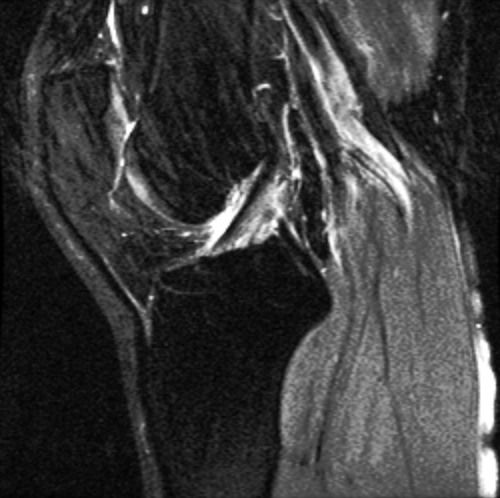}
    \includegraphics[width=0.2\textwidth]{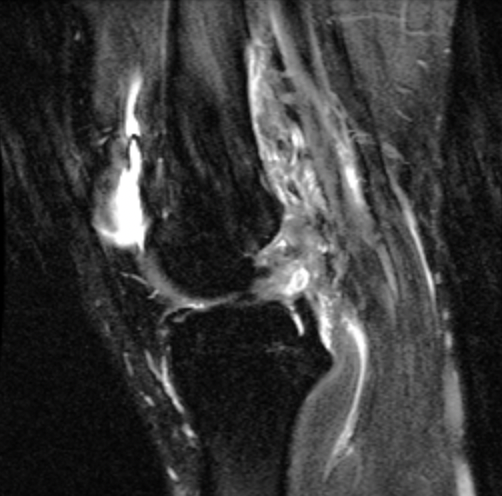}
    \caption{Instances of normal (left) and meniscus tear (right) from the MRNet dataset for our downstream ACL meniscus tear detection task.}
    \label{fig:acl_examples}
\end{figure}

As described in \autoref{sec:intro}, we used two fine-tuning datasets for our downstream tasks: the MRNet dataset of 1021 ACL knee MRI exams \cite{StanfordML2023MRNet}, and the Kaggle Breast Ultrasound Image dataset including 780 US exams \cite{Shah2021BreastUltrasound}. All images are $256 \times 256$. Examples from each dataset are shown in \autoref{fig:breast_ultrasound_examples} and \autoref{fig:acl_examples}.

\subsection{Preprocessing}


We applied various image augmentations for training data preprocessing: random rotations ($\pm10^\circ$), horizontal and vertical shifts (up to 10\%), shear (10\%), zoom (10\%), and horizontal flips. We also shuffled the training dataloader to prevent overfitting to image indices. For all images, we employed 'Caffe'-style preprocessing: RGB to BGR conversion and normalization using ImageNet statistics. Our experiments validated that these steps enhanced model training by stabilizing gradients and reducing sensitivity to input variations (see \autoref{sec:additional_image_prep}).

 \subsection{Splits}

 The original RadImageNet paper used five fold cross-validation during training, which we mimicked for our baseline and removed for all other experiments. We concatenated all data together, modified the split distribution to $75$\% train, $15\%$ validation, and $10\%$ test, and employed stratification on target class to ensure balanced target distributions across splits. This split yields $585/117/78$ images for the breast task and $765/153/103$ images for the ACL task. The increased validation split size and stratification proved highly beneficial to model checkpointing stability, with fewer bad local minima (as tracked by a training monitor script to detect majority class guessing), and smoother validation AUC curves during training.


\section{Experiments and Results}



Our primary quantitative metric is AUC, though we also record F1 and accuracy (see \autoref{sec:bestmodels} and \autoref{subsec:additional_grid_res}).

\subsection{Large Grid Search}

We began our search for optimal breast and ACL classification models with a large grid search framework, searching over a reasonable space of architecture and hyperparameter configurations. Though our compute capabilities limited us to $176$ experiments, we were careful to select a grid search parameter space that was nearby to best parameter choices indicated by initial random training experiments. All experiments in this large grid search were limited to $5$-epochs, as we found $5$ epochs to be sufficient to identify promising experiments in initial testing, and longer training was computationally unviable on our project timeline.

For all grid search trials (\autoref{tab:grid_params}), we standardized several parameters: initial learning rate (\texttt{1e-4}), batch size (\texttt{64}), image size (\texttt{256}), total epochs (\texttt{5}), freeze configuration (\texttt{freezeall}), LR decay factor (\texttt{0.5}), dropout probability (\texttt{0.5}), kernel size for convolutional layers (\texttt{2}), AMP enabled (\texttt{True}), and disabled cross-validation folds (\texttt{False}).

\begin{table}[h]
\centering
\caption{Experiment Hyperparameters and Architecture Variations}
\label{tab:grid_params}
\begin{tabular}{@{}p{1.5cm}p{6cm}@{}}
\toprule
\textbf{Category} & \textbf{Options} \\
\midrule
\textbf{Tasks} & \texttt{["breast", "acl"]} \\
\textbf{Pretrain} & \texttt{["ImageNet", "RadImageNet"]} \\
\textbf{Backbones} & \texttt{["ResNet50", "DenseNet121"]} \\
\textbf{Classifiers} & \texttt{["Linear", "NonLinear", "Conv", "ConvSkip"]} \\
\midrule
\textbf{LR Decay} & \texttt{["beta", "cosine"]} \\
\textbf{FC Ratios} & \texttt{[0.5, 1.0]} \\
\textbf{Num Filters} & \texttt{[4, 16]} \\
\bottomrule
\end{tabular}
\footnotesize
\\[10pt]
"Pretrain" refers to the pre-training weights loaded into the backbone model. "LR Decay" refers to the learning rate decay method iterated each epoch. "FC Ratios" refers to the ratio in size of hidden layer to feature extraction layer (final backbone layer) in relevant classifiers (NonLinear, Conv, ConvSkip). "Num Filters" refers to the number of filters used in convolutional layers for relevant classifiers (Conv, ConvSkip). 
\end{table}

\begin{figure}[htbp]
    \centering
    \includegraphics[width=0.4\textwidth]{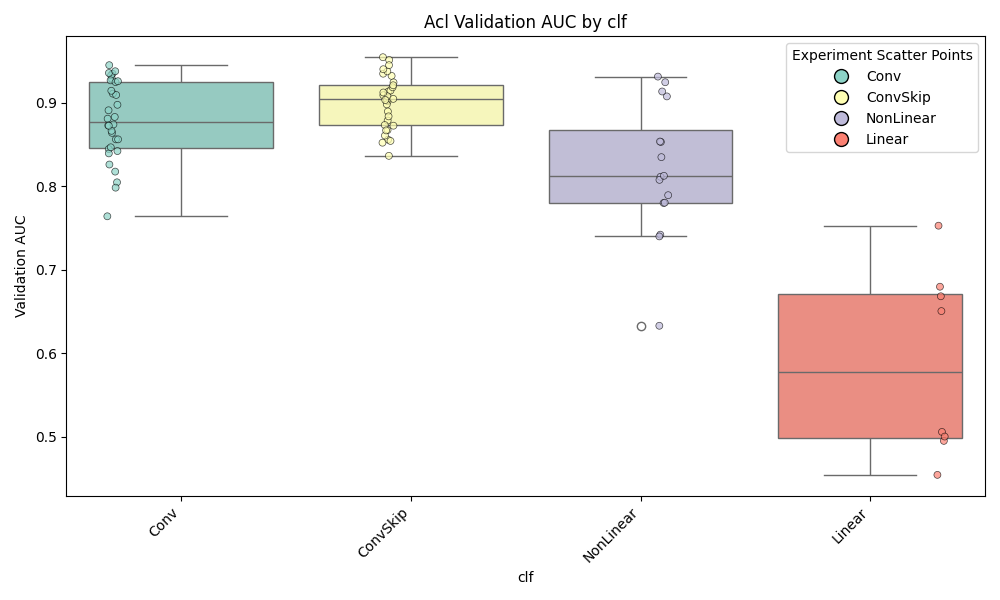}
    \caption{ConvSkip yields better average validation AUCs overall.}
    \label{fig:overall_val_auc_clf}
    \end{figure}

Using validation AUC distributions as a decision-maker, our grid search results suggest that 
\texttt{ResNet50} (see \autoref{fig:overall_val_auc_backbone}) with \texttt{ConvSkip} classifier architecture (see \autoref{fig:overall_val_auc_clf}), \texttt{cosine} annealing learning rate decay (see \autoref{fig:overall_val_auc_lrdecay}), and $16$ convolutional filters (see \autoref{fig:overall_val_auc_numfilters}) perform best.


\begin{figure}[htbp]
    \centering
    \includegraphics[width=0.4\textwidth]{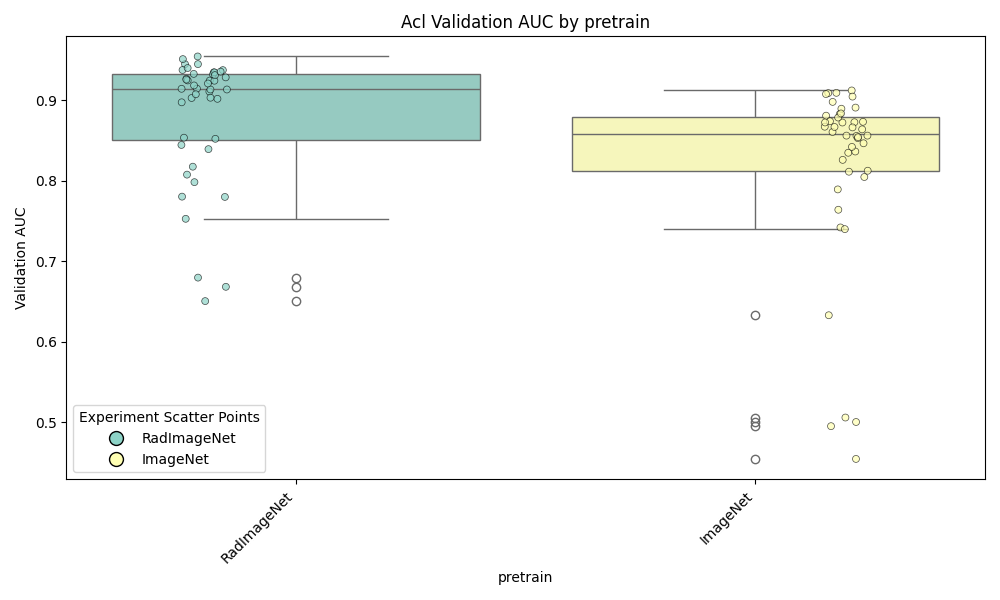}
    \includegraphics[width=0.4\textwidth]
    {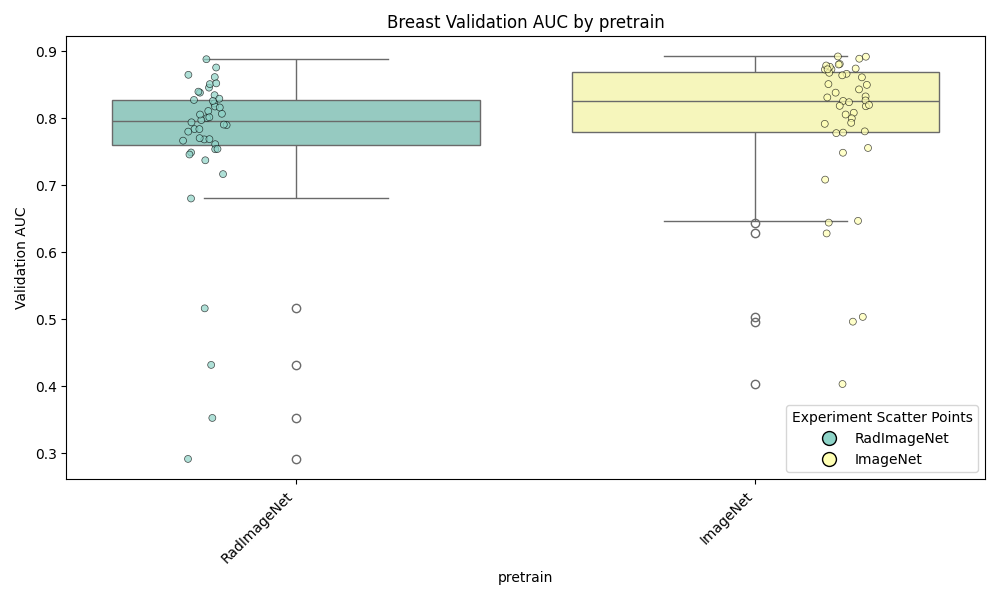}
    \caption{RadImageNet pre-training benefits ACL task, but ImageNet pre-training is superior for Breast task.}
    \label{fig:val_auc_pretrain}
    \end{figure}

Interestingly, we find that \texttt{ImageNet} pre-training weights perform better for the breast cancer task, whereas \texttt{RadImageNet} pre-training weights perform better for the ACL tear task (see \autoref{fig:val_auc_pretrain}).
    
    \subsubsection{Large Grid Search Statistical Analysis }

    \label{subsec:largegridsearch}

Due to the interconnectedness of data points in our grid search, traditional statistical tests like Welch’s t-test or Wilcoxon Rank Sum test may not fully account for the true variability in our study, as these tests assume data independence. The positive correlation among observations could lead to an underestimation of variance. When models are uniformly fine-tuned and evaluated using the same or overlapping datasets, observed performance variances not only reflect the influence of varied hyperparameters but also encompass inherent data noise, which itself is not independent across different model evaluations. With these considerations in mind, we conducted some preliminary Wilcoxon Rank Sum tests which can be found in \autoref{subsec:wilcoxon}.


To robustly estimate the effects of pretraining data and classifier architecture on test AUC, we employed mixed-effects models. These models distinguish between fixed effects, which represent our primary variables of interest, and random effects, which account for variations due to other hyperparameters and specific fine-tuning tasks. This structured approach enhances our capacity to isolate the effects of distinct model features on performance.

Our first mixed-effects model treats pretraining data as the fixed effect. The analysis reveals a positive, though not statistically significant, association with test AUC for models pretrained with RadImageNet (results in \autoref{tab:compact_lmm_pretraining} and more detail in \autoref{tab:full_lmm_pretraining}). This outcome is consistent with our Wilcoxon test results, which highlight RadImageNet's relative effectiveness in ACL task performance compared to ImageNet, particularly as both tasks involve the same number of observations (88 each). This parallel evaluation helps to ensure that our findings are not confounded by differences in sample size.

    \begin{table}[!htbp] \centering 
      \caption{Linear Mixed Model Results for Pretraining} 
      \label{tab:compact_lmm_pretraining}
    \begin{tabular}{@{\extracolsep{5pt}}lcc} 
    \\[-1.8ex]\hline 
    \hline \\[-1.8ex] 
     & Estimate & Pr($>|t|$) \\ 
    \hline \\[-1.8ex] 
    Intercept & 0.738 & 0.0006*** \\ 
    RadImageNet & 0.016 & 0.0543. \\ 
    \hline \\[-1.8ex] 
    Observations & \multicolumn{2}{c}{176} \\ 
    \hline 
    \hline \\[-1.8ex] 
    \end{tabular} 
    \end{table}

In our second mixed-effects model, we focus on the impact of different classifiers with 'Conv' set as the baseline. This analysis indicates that 'Linear' and 'NonLinear' classifiers significantly underperform. In contrast, the 'ConvSkip' classifier shows a promising improvement, though it does not achieve statistical significance at the 5\% level (see \autoref{tab:compact_lmm_clf} and \autoref{tab:full_lmm_clf} for more detail). These findings corroborate our observational data, suggesting that the 'ConvSkip' configuration consistently yields the most favorable results. Notably, the estimates for classifier choice are larger than pre-training weight choice, suggesting a more pronounced impact of architectural choices over pretraining data.
        
    \begin{table}[!htbp] \centering 
      \caption{Linear Mixed Model Focusing for Classifier} 
      \label{tab:compact_lmm_clf}
    \begin{tabular}{@{\extracolsep{5pt}}lcc} 
    \\[-1.8ex]\hline 
    \hline \\[-1.8ex] 
     & Estimate & Pr($>|t|$) \\ 
    \hline \\[-1.8ex] 
    Intercept & 0.829 & 0.000983*** \\ 
    ConvSkip & 0.018 & 0.0755. \\ 
    Linear & -0.313 & $<$2e-16*** \\
    NonLinear & -0.038 & 0.0036** \\
    \hline \\[-1.8ex] 
    Observations & \multicolumn{2}{c}{176} \\ 
    \hline 
    \hline \\[-1.8ex] 
    \end{tabular} 
    \end{table}

\subsection{Backbone Layer Unfreezing Experiments}

We conducted a focused grid search to determine the optimal number of ResNet50 layer groups to unfreeze for each task, testing configurations unfreezing the top $1$, $3$, $5$, $7$, and $9$ layers, as well as a fully frozen model.The experiments were conducted over 10 epochs, keeping other hyperparameters consistent with the best results from the large grid search (\autoref{subsec:largegridsearch}).

\begin{figure}[htbp]
    \centering
    \includegraphics[width=0.4\textwidth]{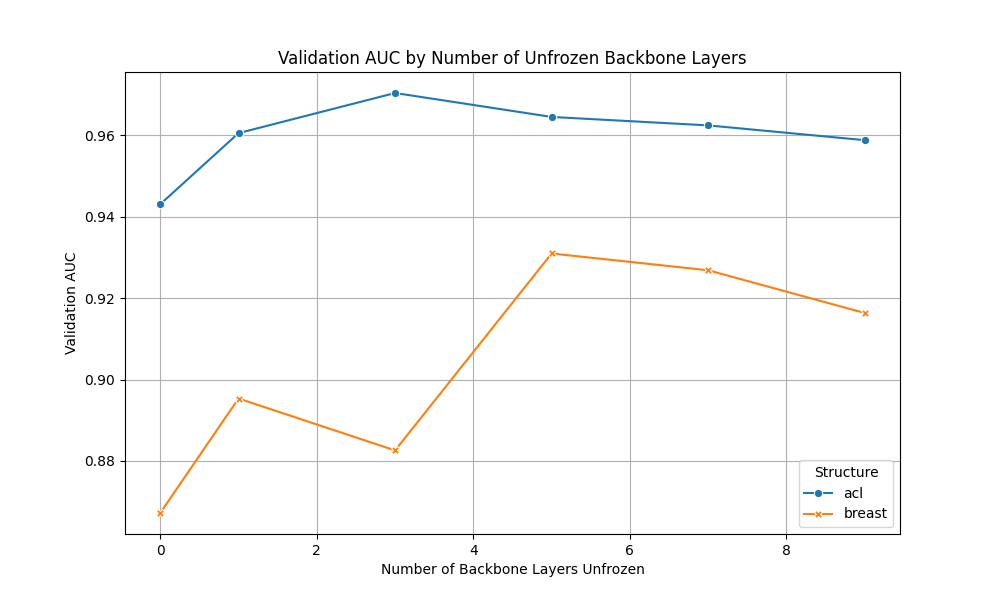}
    \caption{Validation AUC for various amounts of unfreezing, by task.}
    \label{fig:unfreezing}
\end{figure}

We find that the best model performance, as measured by validation AUC, occurs for $3$-$5$ unfrozen \texttt{ResNet50} backbone layer groups (see \autoref{fig:unfreezing}). All unfreezing experiments outperform frozen backbone learning, but there does seem to be a trade-off between increased expressivity and gradient dispersion, whereby too much unfreezing might make learning harder \cite{liu2024fun}. 

\subsection{Final Best Model Selection}

We refined the optimal weight freezing strategy and grid search architecture by tweaking hyperparameters to optimize performance trajectory and learning curve smoothness for the ACL and breast models towards achieving the highest performance possible, which we monitored through TensorBoard. We achieved a large gain in performance for both tasks by using SGD with Nesterov momentum, momentum scalar $0.9$, and a weight decay strength of $0.4$. Best final ACL and breast models were trained for $30$ epochs to get closer to convergence.
As epochs continued, we noted that our breast model was overfitting to the training dataset. To mitigate this, we experimented with various dropout probabilities and optimizers, increased weight decay strength, and re-tested various classifiers, FC hidden layer sizes, and kernel sizes. Ultimately, we trained our best breast model with the architecture in \autoref{tab:architecture_breast_best}.

\begin{table}[htbp]
    \centering
    \caption{Key Hyperparameters for Best Breast Model}
    \label{tab:architecture_breast_best}
    \begin{tabular}{ll}
        \toprule
        \textbf{Hyperparameter} & \textbf{Value} \\
        \midrule
        \texttt{pretrain} & \texttt{ImageNet} \\
        \texttt{backbone\_model\_name} & \texttt{ResNet50} \\
        \texttt{clf} & \texttt{ConvSkip} \\
        \texttt{structure} & \texttt{unfreezetop5} \\
        \texttt{dropout\_prob} & \texttt{0.5} \\
        \texttt{fc\_hidden\_size\_ratio} & \texttt{1.0} \\
        \texttt{num\_filters} & \texttt{16} \\
        \texttt{kernel\_size} & \texttt{2} \\
        \texttt{lr\_decay\_method} & \texttt{cosine} \\
        \texttt{lr} & \texttt{5e-4} \\
        \bottomrule
    \end{tabular}
\end{table}

\begin{figure}[htbp]
    \centering
    \includegraphics[width=0.4\textwidth]{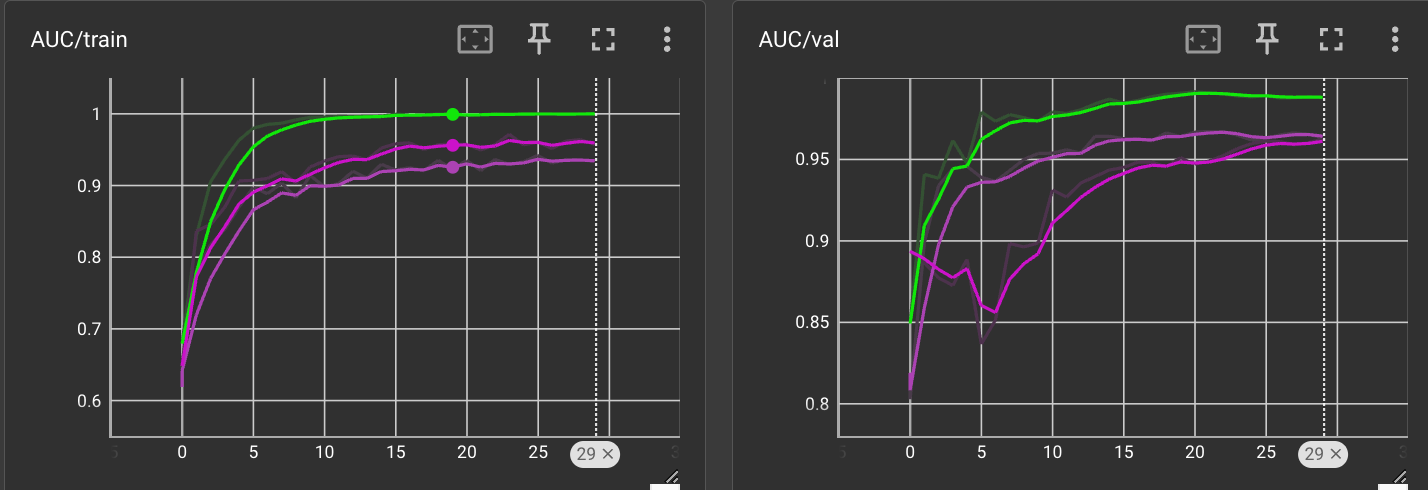}
    \includegraphics[width=0.4\textwidth]{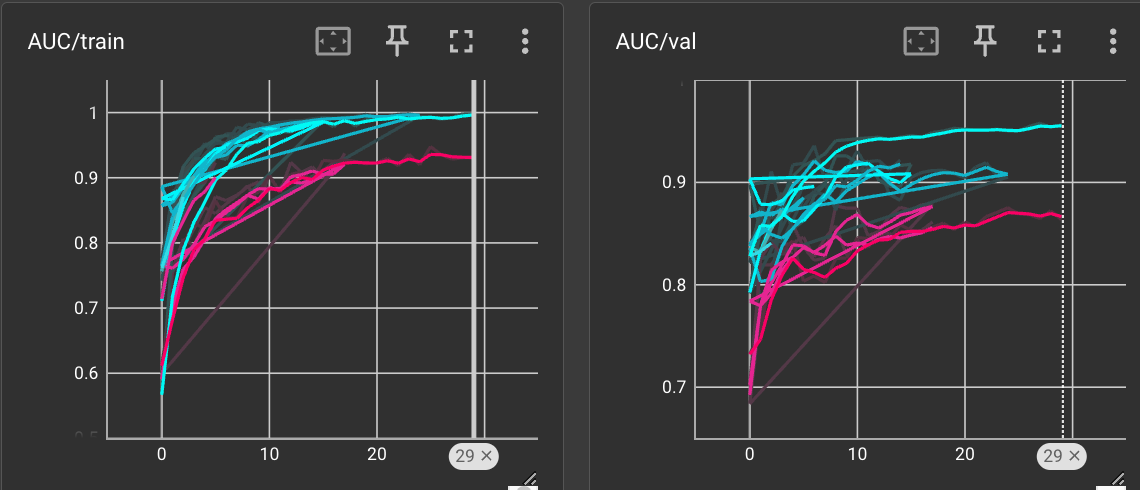}
    \caption{Validation AUC caps out for RadImageNet initialized best models (pink), whereas ImageNet initialized ACL models (green) and ImageNet initialized Breast models (blue) achieve better convergence.}
    \label{fig:radimagenet_caps_out}
\end{figure}

The ACL task proved easier to optimize for, though we hit an upper bound on performance with the \texttt{RadImageNet} weight initialization. Though \texttt{RadImageNet} was more effective at enabling quick ACL model learning during the small $5$ epoch experiments of the large grid search (\autoref{subsec:largegridsearch}), performance plateaued during longer epoch runs across a variety of architectures (see \autoref{fig:radimagenet_caps_out}). As such, we reverted to ImageNet initializations for the ACL task and trained our best ACL model with the architecture detailed in \autoref{tab:architecture_acl_best}.

\begin{table}[htbp]
    \centering
    \caption{Key Hyperparameters for Best ACL Model}
    \label{tab:architecture_acl_best}
    \begin{tabular}{ll}
        \toprule
        \textbf{Hyperparameter} & \textbf{Value} \\
        \midrule
        \texttt{pretrain} & \texttt{ImageNet} \\
        \texttt{backbone\_model\_name} & \texttt{ResNet50} \\
        \texttt{clf} & \texttt{ConvSkip} \\
        \texttt{structure} & \texttt{unfreezetop5} \\
        \texttt{dropout\_prob} & \texttt{0.5} \\
        \texttt{fc\_hidden\_size\_ratio} & \texttt{0.5} \\
        \texttt{num\_filters} & \texttt{16} \\
        \texttt{kernel\_size} & \texttt{4} \\
        \texttt{lr\_decay\_method} & \texttt{cosine} \\
        \texttt{lr} & \texttt{1e-3} \\
        \bottomrule
    \end{tabular}
\end{table}

Our best breast model achieves a \textbf{test AUC of 0.9641} and our best ACL model achieves a \textbf{test AUC of 0.9969}. Compared to a 2021 study evaluating breast lesion malignancy detection performance between machine learning models and radiologists, where the model achieved an AUC of 0.855 and the radiologists achieved an average AUC of 0.805, our breast model seems highly promising \cite{sun2020deep}. Compared with the original RadImageNet paper's reported AUC results of 0.94 ± 0.05 for the breast task and 0.97 ± 0.03 for the ACL task, our best models are competitive. Since Mei et al. (2022) \cite{Mei2022RadImageNet} do not clarify whether reported ranges are empirical or theoretical, nor whether the results are training, validation, or test AUCs, it is challenging to rigorously compare performance. Our full best model performance metrics can be found in \autoref{tab:bestmodels}.

To compare RadImageNet and ImageNet pretraining weights statistically, we used DeLong's tests for correlated ROC curves (\autoref{tab:delong_results}), comparing the best ACL and breast models trained with both ImageNet and RadImageNet. The results indicate that ImageNet's superior performance is statistically significant at the 5\% level for both tasks. Notably, while the AUC difference for the breast task is larger, it demonstrates less significance due to the smaller test set size and greater variability. Additional analysis using McNemar's Chi-squared test can be found in \autoref{subsec:mcnemar} and supports the result that the relationship is much stronger for ACL, as it does not detect a significant difference for the breast task.

\begin{table}[htbp]
\centering
\caption{DeLong's Test: Difference in Best Model Performance Between ImageNet and RadImageNet}
\label{tab:delong_results}
\begin{tabular}{lcc}
\toprule
& \textbf{ACL Task} & \textbf{Breast Task} \\ 
\midrule
\textbf{Z-value} & 2.3914 & 1.974 \\
\textbf{p-value} & 0.01678 & 0.04838 \\
\textbf{95\% CI} & [0.0068, 0.0683] & [0.0005, 0.1498] \\
\midrule
\small\textbf{AUC of IMG} & 0.9969 & 0.9641 \\
\small \textbf{AUC of RAD} & 0.9594 & 0.8889 \\
\bottomrule
\end{tabular}
\end{table}
 
\subsection{Visual Interpretation}

\begin{figure}[htbp]
    \centering
    \includegraphics[width=0.5\textwidth]{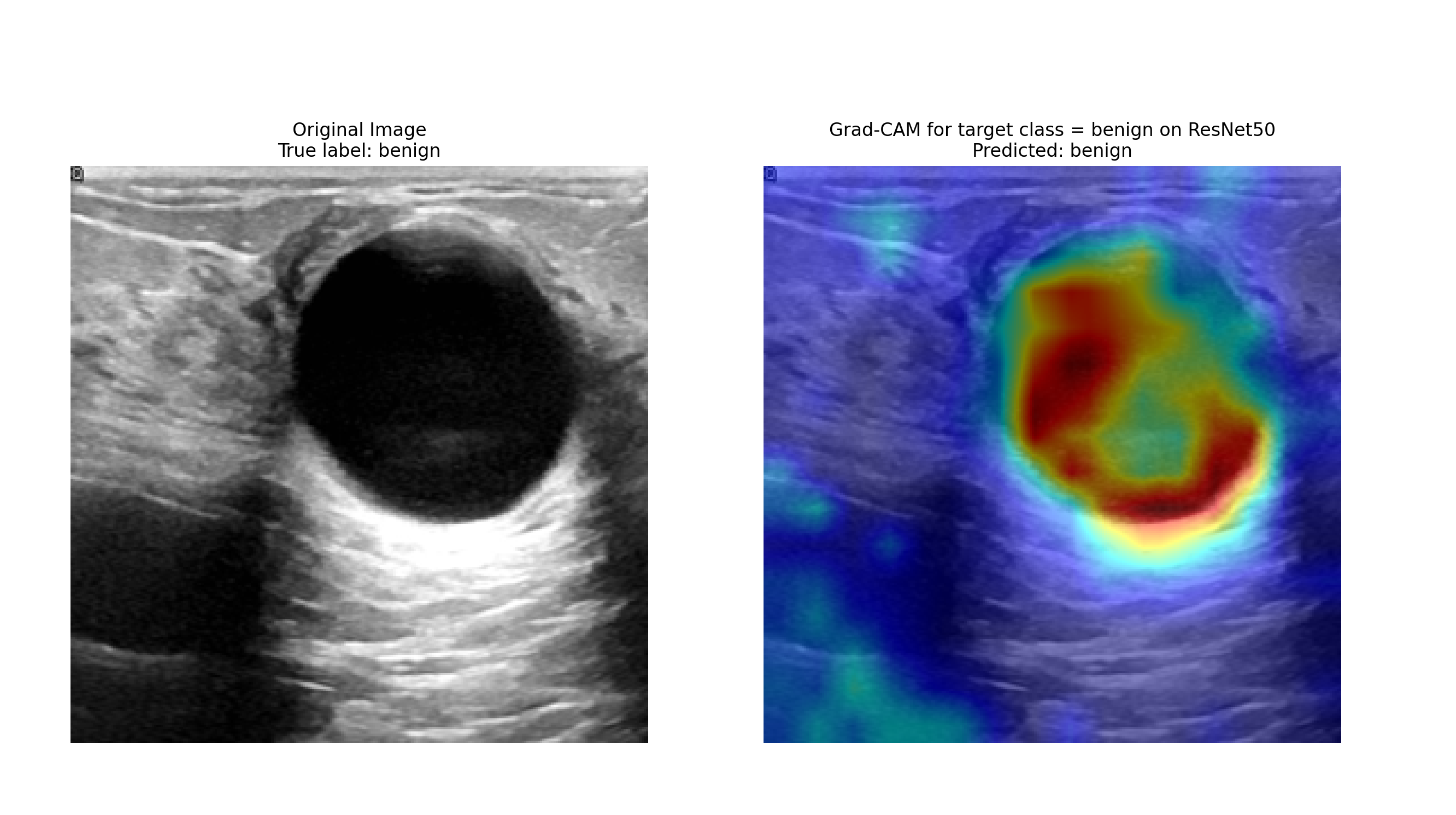}
    \caption{Breast Malignancy Classification model examines a large benign lesion. Hotter color indicates increased pixel importance.}
    \label{fig:big_benign}
\end{figure}

To better understand diagnostic decisions made by our best breast and ACL models on validation data, we use Grad-CAM, as described in \autoref{sec:methods}, to visualize convolutional 'attention' overlayed onto input images alongside the original image, true image label and model prediction. The heatmaps of input region importance reveal highly promising model behavior and good interpretability. In  \autoref{fig:big_benign}, for example, we notice that our best breast model not only learns to weakly segment the breast lesion, but also appears to pay more attention  to the edges of the lesion, which falls in line with clinical best practices. Malignancy is often indicated primarily by the smoothness or roughness of lesion edges \cite{mu2008classification}. 

\vspace{-4mm}

\begin{figure}[htbp]
    \centering
    \includegraphics[width=0.5\textwidth]{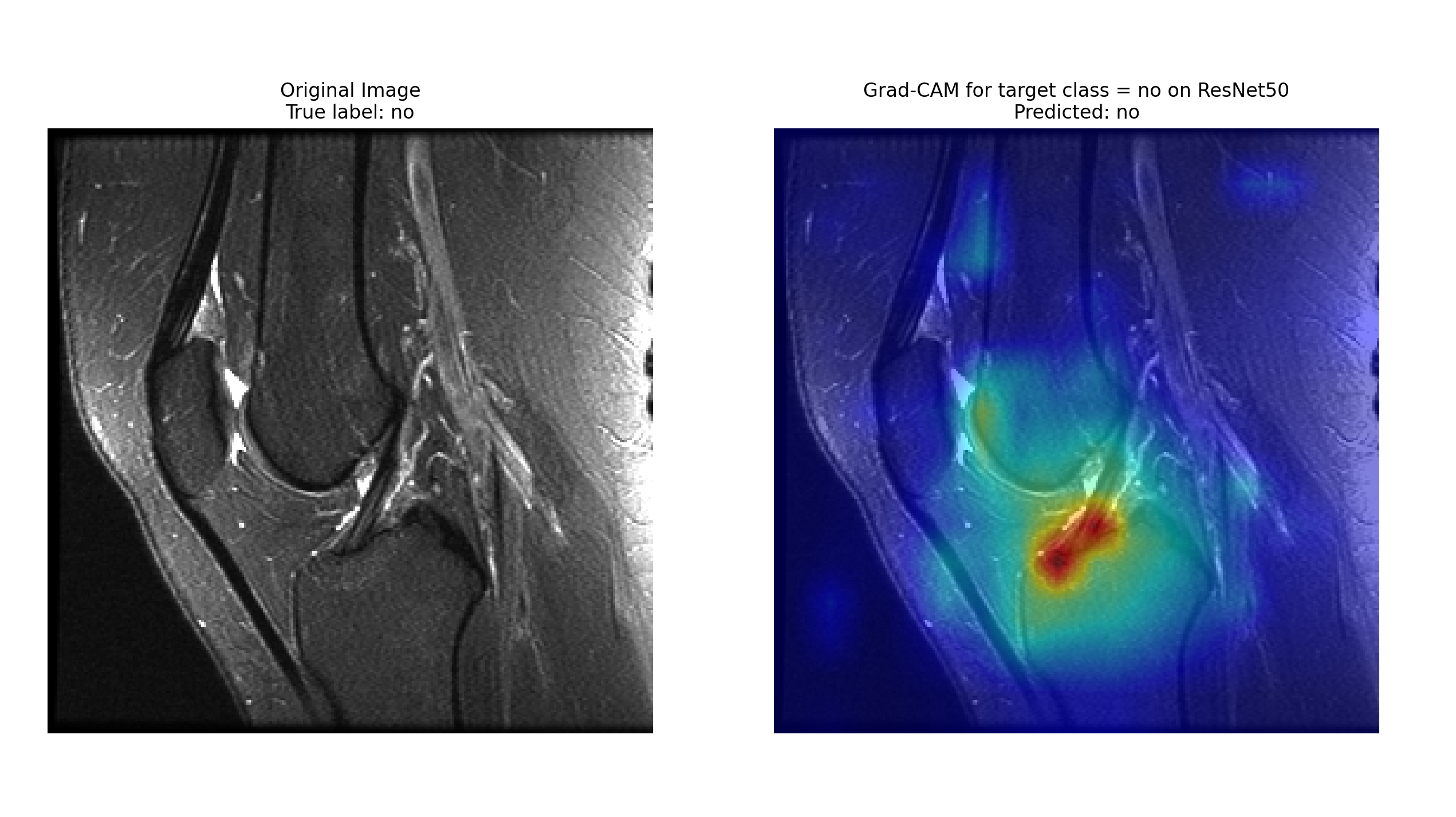}
    \caption{ACL Tear Classification model finds the ACL. Hotter color indicates increased pixel importance.}
    \label{fig:acl_gradcam}
\end{figure}

Though the ACL tear detection task is much easier, we similarly see in \autoref{fig:acl_gradcam} that our model learns to pay attention to the correct region of the input image and does a decent job of locating the ACL, despite having been trained for classification, not segmentation. 

\vspace{-4mm}

\subsubsection{Failure Cases}

\begin{figure}[!htbp]
    \centering
    \includegraphics[width=0.5\textwidth]{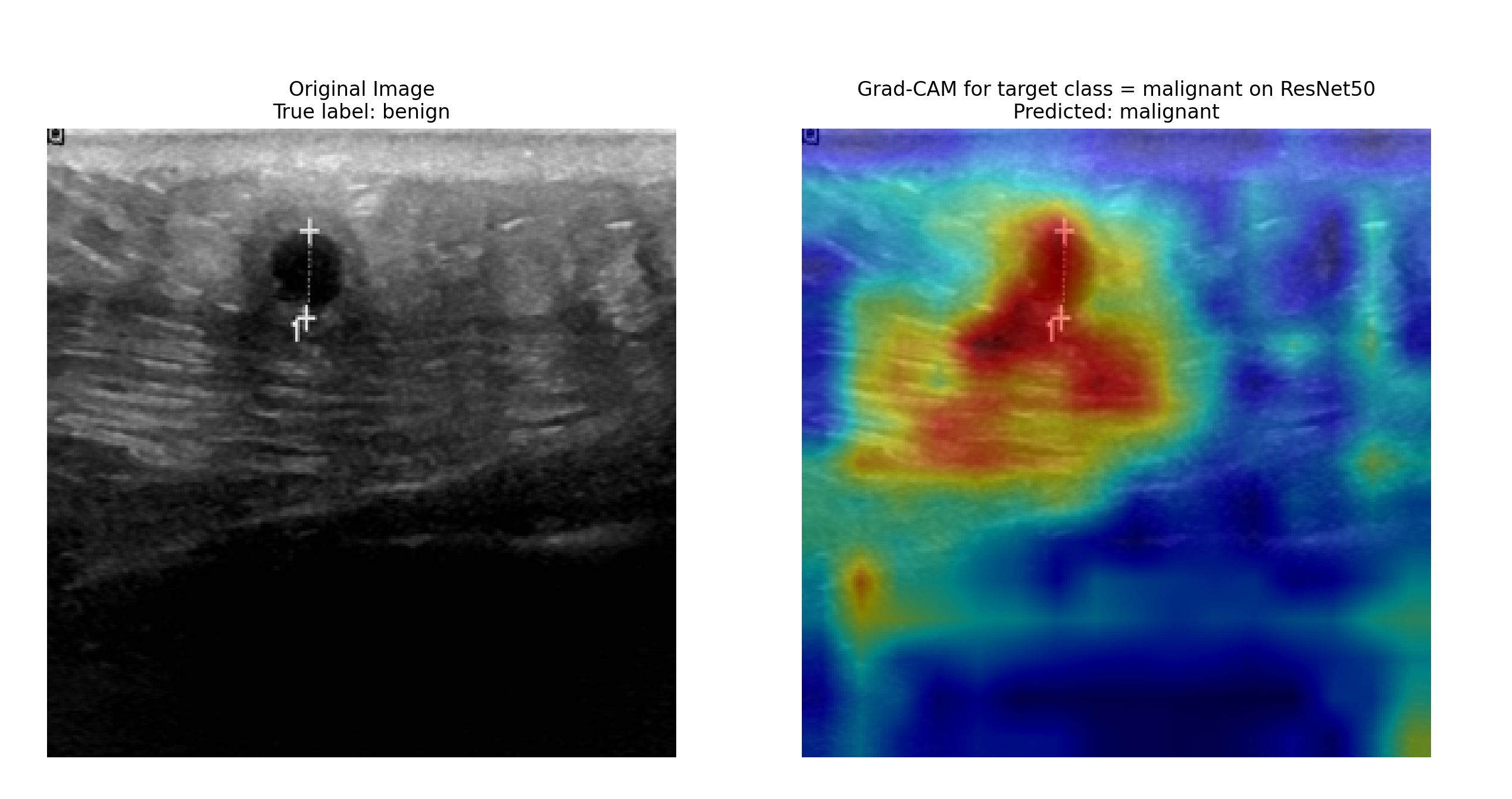}
    \caption{Breast lesion classification mistake.}
    \label{fig:breast_mistake}
\end{figure}

While neither of our models achieves $100\%$ accuracy, a promising observation is that model failures seem to be easily understood via Grad-CAM visual interpretation. \autoref{fig:breast_mistake} is an example where the Breast lesion classification model makes a mistake. It appears that the malignancy prediction arises from the textured shadow regions underneath the benign lesion. The model may predict this lesion to be malignant because it falsely determines those shadows to be indicative of tumor metastasis, a common indication of malignancy. 

\vspace{-2mm}


\section{Conclusion}

We could not replicate Mei et al. (2022)'s findings of superior RadImageNet pre-training performance. Our best breast lesion and ACL tear models demonstrated statistically significant superior performance when pre-trained on ImageNet as opposed to RadImageNet. This divergence was especially notable on the ACL tear task. That said, the ACL tear task demonstrated initial performance gains in early epochs. This same effect was not observed with the breast lesion classification task, which may be attributed to the lack of breast ultrasound images in RadImageNet's pre-training data. This suggests that RadImageNet pre-training benefits might be isolated to the specific anatomical content included during pre-training. Expanding RadImageNet to include a broader spectrum of imaging modalities and anatomical areas could provide significant benefits.
ImageNet's superior performance may also be attributed to discrepancies between RadImageNet PyTorch vs. TensorFlow weights (see \autoref{subsec:tensorflow_inception_better}) or the particular subspace of architectures and hyperparameters we investigate.  Conversely, ImageNet's broader diversity likely aids general performance by enhancing initial layer training on diverse features such as more varied textures, shapes, and scenes. On the other hand, our study underscores that the impact of classifier architecture, optimizer, learning rate and other hyperparameters may exceed the influence of pre-training data (see \autoref{tab:full_lmm_pretraining} and \autoref{tab:full_lmm_clf}).




For future work, combining the strengths of both ImageNet and RadImageNet through combined dataset pre-training or hybrid model architectures could enhance fine-tuned model performance. A simple first approach could be to concatenate the features extracted via RadImageNet and ImageNet and present this full feature set as input into classifier layers. This approach, along with the exploration of adaptive unfreezing techniques such as Liu et al. (2024) \cite{liu2024fun} and the incorporation of newer, larger datasets like the expanded ImageNet is likely to further refine performance of the breast lesion and ACL tear models presented here.

\clearpage

\section*{Supplementary Material}


All code for this project can be found at \url{https://github.com/danielfrees/bacon}.

\section{Contributions \& Acknowledgements }

\begin{itemize}

\item The original TensorFlow code for RadImageNet served as inspiration for this project: \url{https://github.com/BMEII-AI/RadImageNet}

\item \textbf{\textit{Daniel:} }Debugged TensorFlow code, refactored PyTorch code, debugged partial backbone unfreezing,  implemented \texttt{Conv} classifier, architecture/hyperparameter/unfreezing grid search \& visualizations, Grad-CAM \& heatmaps, TensorBoarding, and other improvements such as learning rate decay methods, data splitting, Caffe preprocessing, optimizer testing, additional metrics (F1, acc), etc. Also implemented code linting, pytesting, etc. for the repo wrapped with a pre-commit config.

\item \textbf{\textit{Moritz:}} Implemented original PyTorch code (translated Mei et al.'s TensorFlow architecture), added ability to partially unfreeze layers, devised and performed all statistical analysis, created flow charts for model pretraining/finetuning/inference visualization

\item \textbf{\textit{Aditri:}} Implemented Linear, NonLinear and ConvSkip model classifiers. Completed data pre-processing and data augmentation experiments with color jitter, affine transformations, flips, scaling, etc. Literature review for extensions beyond baselines.

\end{itemize}

{\small
\bibliographystyle{ieee}
\bibliography{cs231report}

\begin{thebibliography}{10}\itemsep=-1pt

\bibitem{imagenet_dataset}
Imagenet dataset - sciencedirect topics, 2023.
\newblock [Online; accessed 5-June-2024].

\bibitem{tensorflow2015-whitepaper}
M.~Abadi, A.~Agarwal, P.~Barham, E.~Brevdo, Z.~Chen, C.~Citro, G.~S. Corrado, A.~Davis, J.~Dean, M.~Devin, S.~Ghemawat, I.~Goodfellow, A.~Harp, G.~Irving, M.~Isard, Y.~Jia, R.~Jozefowicz, L.~Kaiser, M.~Kudlur, J.~Levenberg, D.~Man\'{e}, R.~Monga, S.~Moore, D.~Murray, C.~Olah, M.~Schuster, J.~Shlens, B.~Steiner, I.~Sutskever, K.~Talwar, P.~Tucker, V.~Vanhoucke, V.~Vasudevan, F.~Vi\'{e}gas, O.~Vinyals, P.~Warden, M.~Wattenberg, M.~Wicke, Y.~Yu, and X.~Zheng.
\newblock {TensorFlow}: Large-scale machine learning on heterogeneous systems, 2015.
\newblock Software available from tensorflow.org.

\bibitem{Bates2015}
D.~Bates, M.~M{\"a}chler, B.~Bolker, and S.~Walker.
\newblock Fitting linear mixed-effects models using {lme4}.
\newblock {\em Journal of Statistical Software}, 67(1):1--48, 2015.

\bibitem{RadImageNetGitHub}
{BMEII-AI}.
\newblock Radimagenet: A radiology-focused image dataset for machine learning.
\newblock \url{https://github.com/BMEII-AI/RadImageNet}, 2024.

\bibitem{bressem2020comparing}
K.~K. Bressem, L.~C. Adams, C.~Erxleben, B.~Hamm, S.~M. Niehues, and J.~L. Vahldiek.
\newblock Comparing different deep learning architectures for classification of chest radiographs.
\newblock {\em Scientific Reports}, 10:13590, 2020.

\bibitem{DeLong1988}
E.~R. DeLong et~al.
\newblock Comparing the areas under two or more correlated receiver operating characteristic curves: A nonparametric approach.
\newblock {\em Biometrics}, 44(3):837--845, 1988.
\newblock Accessed 2 June 2024.

\bibitem{deng2009imagenet}
J.~Deng, W.~Dong, R.~Socher, L.-J. Li, K.~Li, and L.~Fei-Fei.
\newblock Imagenet: A large-scale hierarchical image database.
\newblock In {\em 2009 IEEE conference on computer vision and pattern recognition}, pages 248--255. IEEE, 2009.

\bibitem{Efron1979}
B.~Efron.
\newblock Bootstrap methods: Another look at the jackknife.
\newblock {\em Ann. Statist.}, 7(1):1--26, January 1979.

\bibitem{He2015}
K.~He, X.~Zhang, S.~Ren, and J.~Sun.
\newblock Delving deep into rectifiers: Surpassing human-level performance on imagenet classification.
\newblock {\em arXiv preprint arXiv:1502.01852}, 2015.

\bibitem{Kim2022}
H.~E. Kim, A.~Cosa-Linan, N.~Santhanam, M.~Jannesari, M.~E. Maros, and T.~Ganslandt.
\newblock Transfer learning for medical image classification: a literature review.
\newblock {\em BMC Medical Imaging}, 22(69), 2022.

\bibitem{liu2024fun}
C.~C. Liu, J.~Pfeiffer, I.~Vulic, and I.~Gurevych.
\newblock Fun with fisher: Improving generalization of adapter-based cross-lingual transfer with scheduled unfreezing.
\newblock In {\em Proceedings of the North American Chapter of the Association for Computational Linguistics (NAACL)}, Darmstadt, Germany; London, UK; Cambridge, UK, 2024. Ubiquitous Knowledge Processing Lab, Department of Computer Science and Hessian Center for AI (hessian.AI), Technical University of Darmstadt; Google DeepMind; Language Technology Lab, University of Cambridge.

\bibitem{mathivanan2024employing}
S.~K. Mathivanan, S.~Sonaimuthu, S.~Murugesan, H.~Rajadurai, B.~D. Shivahare, and M.~A. Shah.
\newblock Employing deep learning and transfer learning for accurate brain tumor detection.
\newblock {\em Sci Rep}, 14(1):7232, Mar 2024.

\bibitem{mcnemar1947}
Q.~McNemar.
\newblock Note on the sampling error of the difference between correlated proportions or percentages.
\newblock {\em Psychometrika}, 12(2):153--157, Jun 1947.

\bibitem{Mei2022RadImageNet}
X.~Mei, Z.~Liu, P.~M. Robson, B.~Marinelli, M.~Huang, A.~Doshi, A.~Jacobi, C.~Cao, K.~E. Link, T.~Yang, Y.~Wang, H.~Greenspan, T.~Deyer, Z.~A. Fayad, and Y.~Yang.
\newblock Radimagenet: An open radiologic deep learning research dataset for effective transfer learning.
\newblock {\em Radiology: Artificial Intelligence}, 4(5):e210315, 2022.

\bibitem{mu2008classification}
T.~Mu, A.~K. Nandi, and R.~M. Rangayyan.
\newblock Classification of breast masses using selected shape, edge-sharpness, and texture features with linear and kernel-based classifiers.
\newblock 2008.

\bibitem{Panwar2020}
H.~Panwar, P.~Gupta, M.~K. Siddiqui, R.~Morales-Menendez, P.~Bhardwaj, and V.~Singh.
\newblock A deep learning and grad-cam based color visualization approach for fast detection of covid-19 cases using chest x-ray and ct-scan images.
\newblock {\em Journal of Biomedical Informatics}, 2020.

\bibitem{paszke2019pytorch}
A.~Paszke, S.~Gross, F.~Massa, A.~Lerer, J.~Bradbury, G.~Chanan, T.~Killeen, Z.~Lin, N.~Gimelshein, L.~Antiga, A.~Desmaison, A.~K\"{o}pf, E.~Yang, Z.~DeVito, M.~Raison, A.~Tejani, S.~Chilamkurthy, B.~Steiner, L.~Fang, J.~Bai, and S.~Chintala.
\newblock Pytorch: An imperative style, high-performance deep learning library.
\newblock {\em Advances in Neural Information Processing Systems 32}, 2019.
\newblock Software available from pytorch.org.

\bibitem{R2024}
{R Core Team}.
\newblock {\em R: A Language and Environment for Statistical Computing}.
\newblock R Foundation for Statistical Computing, Vienna, Austria, 2021.

\bibitem{pROC2011}
X.~Robin, N.~Turck, A.~Hainard, N.~Tiberti, F.~Lisacek, J.-C. Sanchez, and M.~Müller.
\newblock proc: an open-source package for r and s+ to analyze and compare roc curves.
\newblock {\em BMC Bioinformatics}, 12:77, 2011.

\bibitem{Rose1996}
N.~E. Rose and S.~M. Gold.
\newblock A comparison of accuracy between clinical examination and magnetic resonance imaging in the diagnosis of meniscal and anterior cruciate ligament tears.
\newblock {\em Arthroscopy}, 12(4):398--405, 1996.

\bibitem{selvaraju2019grad}
R.~R. Selvaraju, M.~Cogswell, A.~Das, R.~Vedantam, D.~Parikh, and D.~Batra.
\newblock Grad-cam: Visual explanations from deep networks via gradient-based localization.
\newblock {\em arXiv preprint arXiv:1610.02391}, 2019.

\bibitem{Shah2021BreastUltrasound}
A.~Shah.
\newblock {Breast Ultrasound Images Dataset}.
\newblock \url{https://www.kaggle.com/datasets/aryashah2k/breast-ultrasound-images-dataset}, 2021.

\bibitem{StanfordML2023MRNet}
{Stanford ML Group}.
\newblock {MRNet Competition}.
\newblock \url{https://stanfordmlgroup.github.io/competitions/mrnet/}, 2023.

\bibitem{sun2020deep}
Y.~Sun, Y.~Qu, D.~Wang, Y.~Li, L.~Ye, J.~Du, B.~Xu, B.~Li, X.~Li, K.~Zhang, et~al.
\newblock Deep learning model improves radiologists’ performance in detection and classification of breast lesions.
\newblock 2021.
\newblock * These authors contributed equally to this work.

\bibitem{Wilcoxon1945}
F.~Wilcoxon.
\newblock Individual comparisons by ranking methods.
\newblock {\em Biometrics Bulletin}, 1(6):80--83, 1945.
\newblock Accessed 5 June 2024.

\bibitem{zhang2020}
Q.~Zhang, Y.~N. Wu, and S.-C. Zhu.
\newblock Interpretable convolutional neural networks.
\newblock In {\em Proceedings of the IEEE Conference on Computer Vision and Pattern Recognition (CVPR)}. IEEE, 2020.

\bibitem{Zhao2023}
Z.~Zhao, L.~Alzubaidi, J.~Zhang, Y.~Duan, and Y.~Gu.
\newblock A comparison review of transfer learning and self-supervised learning: Definitions, applications, advantages and limitations.
\newblock {\em Expert Systems with Applications}, 2023.

\end{thebibliography}
}

\newpage
\appendix
\onecolumn
\section*{Appendix}
\label{app}

\section{Full Best Model Results}
\label{sec:bestmodels}

\begin{table*}[htbp]
    \centering
    \caption{Best Models Performance Metrics}
    \label{tab:bestmodels}
    \begin{tabular}{llcccc}
        \toprule
        \textbf{Model} & \textbf{Data Split} & \textbf{Loss} & \textbf{AUC} & \textbf{F1 Score} & \textbf{Accuracy} \\
        \midrule
        \multirow{3}{*}{Breast}
        & Training   & 0.0535 & 0.9981 & 0.9655 & 0.9812 \\
        & Validation & 0.3887 & 0.9561 & 0.7826 & 0.8718 \\
        & Test       & \textbf{0.2250} & \textbf{0.9641} & \textbf{0.8293} & \textbf{0.9103} \\
        \midrule
        \multirow{3}{*}{Breast RadImageNet}
        & Training   & 0.2902 & 0.9304 & 0.7639 & 0.8838 \\
        & Validation & 0.3957 & 0.8635 & 0.6129 & 0.7949 \\
        & Test       & 0.3363 & 0.8889 & 0.7317 & 0.8590 \\
        \midrule
        \multirow{3}{*}{ACL}
        & Training   & 0.0123 & 0.9999 & 0.9965 & 0.9961 \\
        & Validation & 0.1972 & 0.9881 & 0.9383 & 0.9346 \\
        & Test       & \textbf{0.0996} & \textbf{0.9969} & \textbf{0.9739} & \textbf{0.9709} \\
        \midrule
        \multirow{3}{*}{ACL RadImageNet}
        & Training   & 0.2837 & 0.9620 & 0.9102 & 0.8993 \\
        & Validation & 0.3295 & 0.9590 & 0.8553 & 0.8562 \\
        & Test       & 0.3295 & 0.9594 & 0.8350 & 0.8350 \\
        \bottomrule
    \end{tabular}
\end{table*}

\clearpage

\section{Overall Transfer-learning Architecture}
\label{sec:transfer_architecture}

The RadImageNet-based model architecture begins with pre-trained ResNet50, InceptionV3, and DenseNet121 with data from ImageNet or RadImageNet. The resulting model is fine-tuned with data of the same type as the desired classification task. For classification, the pre-trained model can be combined with different classifier layers, such as linear, non-linear, convolutional, and convolutional with skip connections. 
    \begin{figure}[h]
        \centering
        \includegraphics[scale=2]{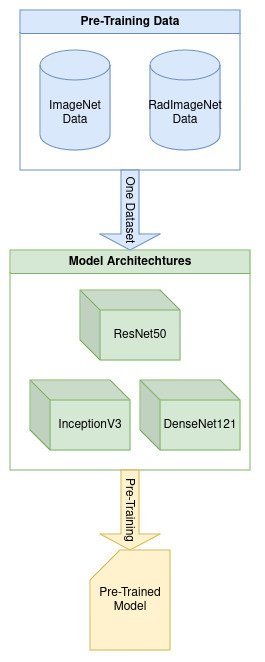}
        \caption{Pre-Training Process}
        \label{fig:enter-label}
    \end{figure}
    \begin{figure}[h]
        \centering
        \includegraphics[scale=2]{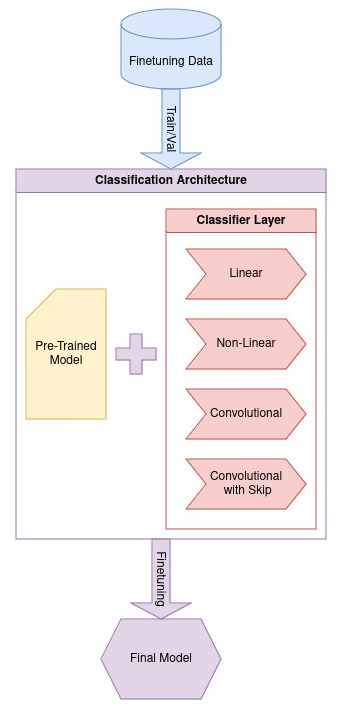}
        \caption{Finetuning Process}
        \label{fig:enter-label}
    \end{figure}
    \begin{figure}[h]
        \centering
        \includegraphics[scale=2]{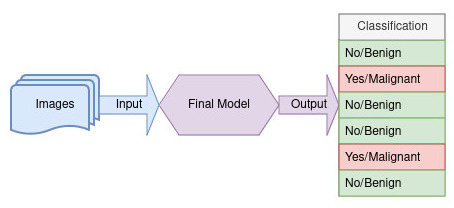}
        \caption{Final Inference}
        \label{fig:enter-label}
    \end{figure}

\clearpage

\section{Additional Grad-CAM visualizations}
\label{sec:more_gradcam}

First, we analyze example mistakes to qualitatively understand the failure points of our best breast and ACL models. 

\begin{figure}[!htbp]
    \centering
    \includegraphics[width=0.8\textwidth]{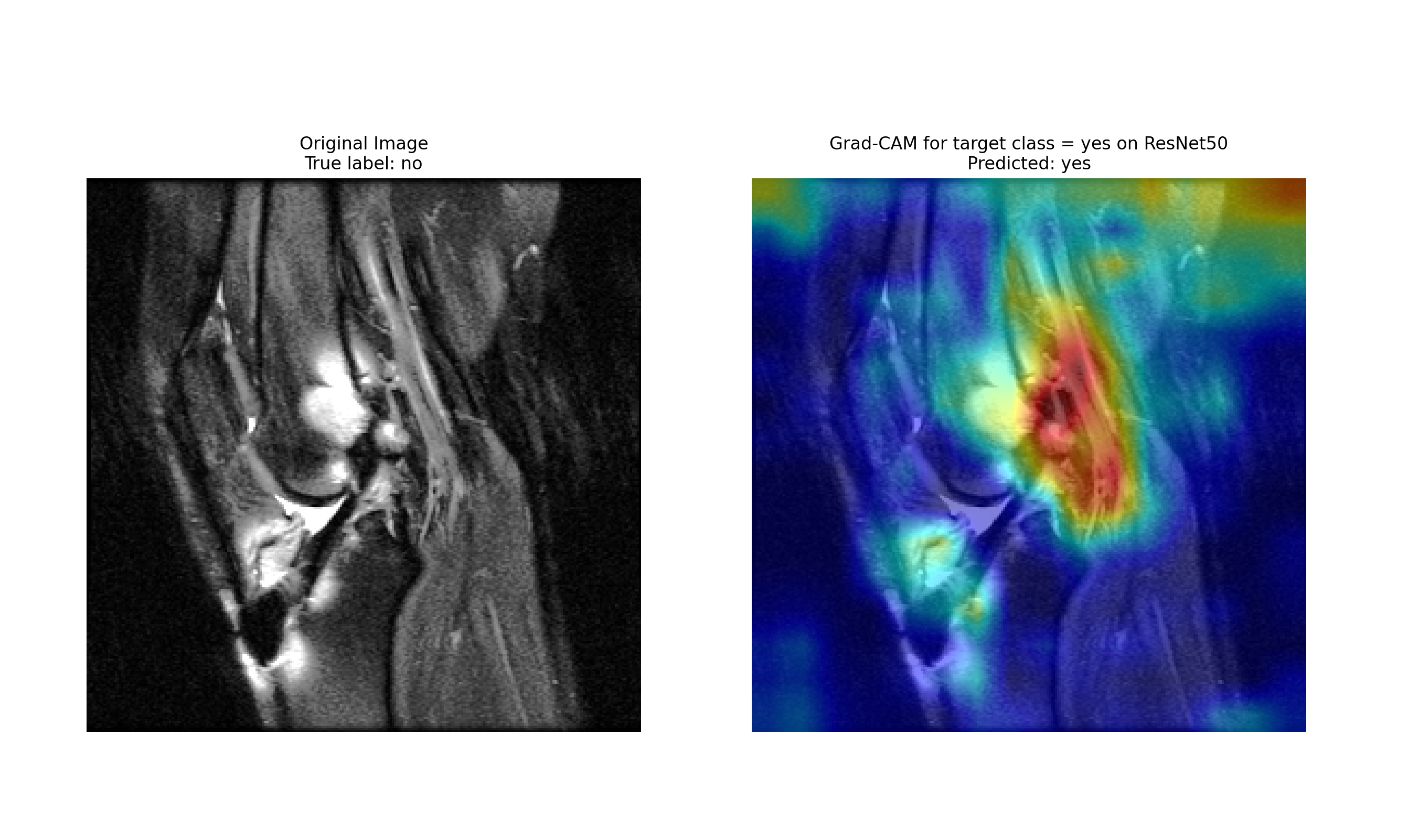}
    \caption{ACL Tear Classification model mistake. }
    \label{fig:acl_gradcam_mistake2}
\end{figure}

First, we see in \autoref{fig:acl_gradcam_mistake2} that the ACL tear model falsely determines this MRI scan patient to have suffered an ACL tear. The input image has some visual artifacts, with 'spotlight' lighting bubbles around the important central regions of the image. Furthermore, this input image contains significantly more visible bone and muscle detail compared to most of the ACL data. As a result, it seems that while the ACL tear model roughly pays attention to the right part of the input, it cannot find an intact ACL (more rigorously, the features extracted by the model which mathematically equate to an intact ACL in the model prediction space are unable to correctly determine that this patient has an intact ACL). 

\clearpage 

Below are a handful of other correct breast classification model predictions visualized with Grad-CAM. 

\begin{figure}[!htbp]
    \centering
    \includegraphics[width=0.9\textwidth]{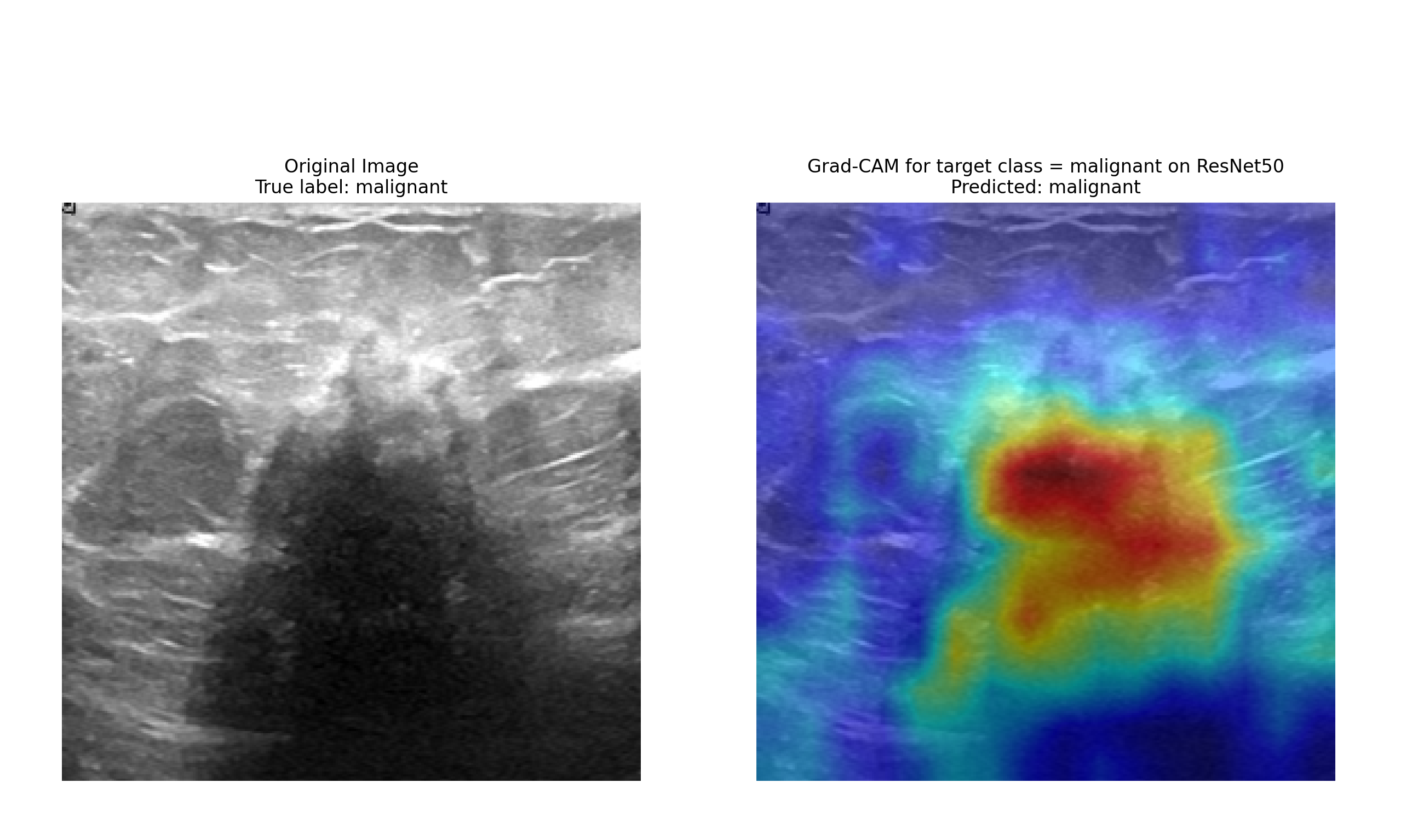}
    \caption{Breast classifier correctly identifies malignancy.}
\end{figure}

\begin{figure}[!htbp]
    \centering
    \includegraphics[width=0.9\textwidth]{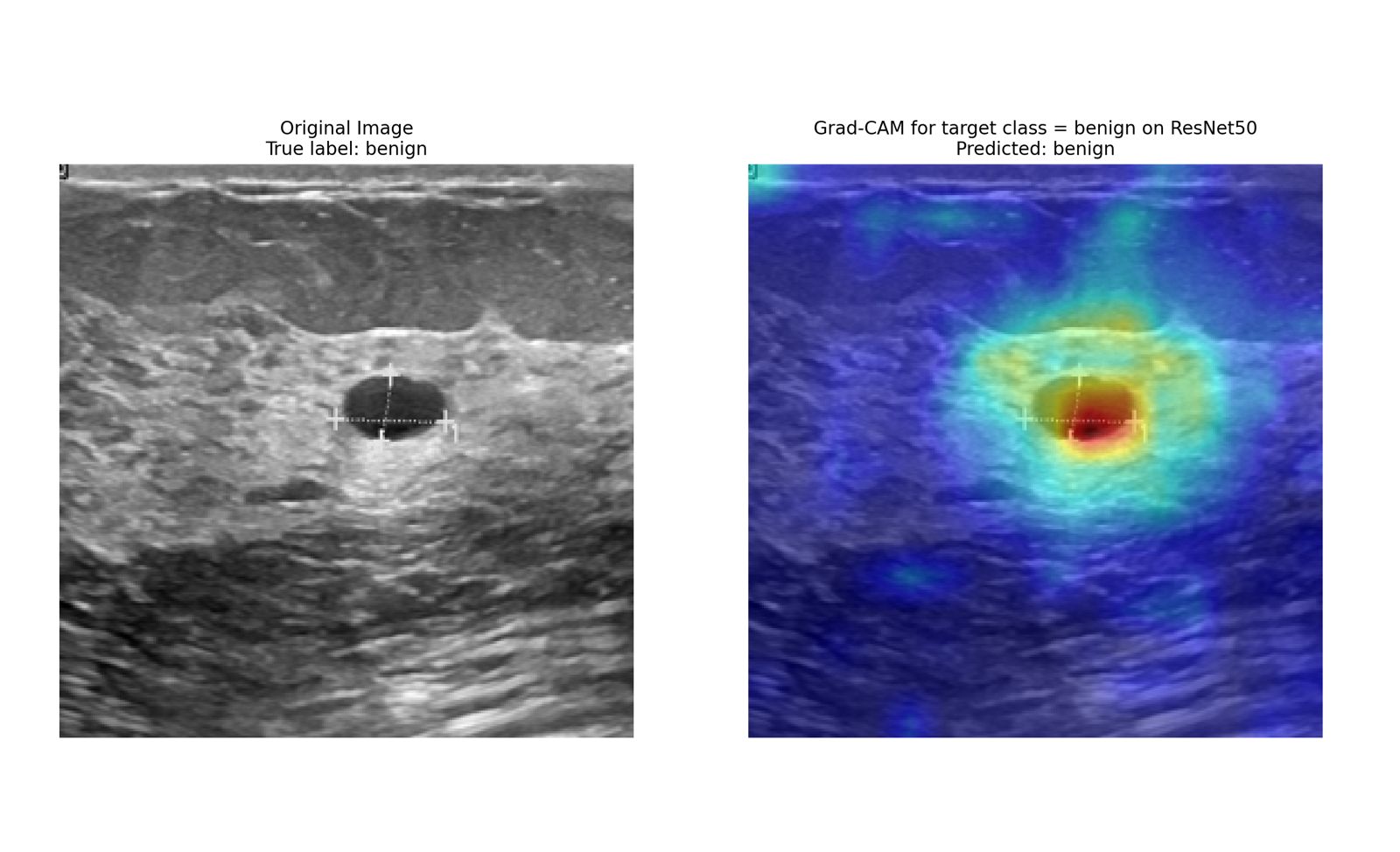}
    \caption{Breast classifier correctly identifies a small benign lesion.}
    
\end{figure}

\begin{figure}[!htbp]
    \centering
    \includegraphics[width=0.9\textwidth]{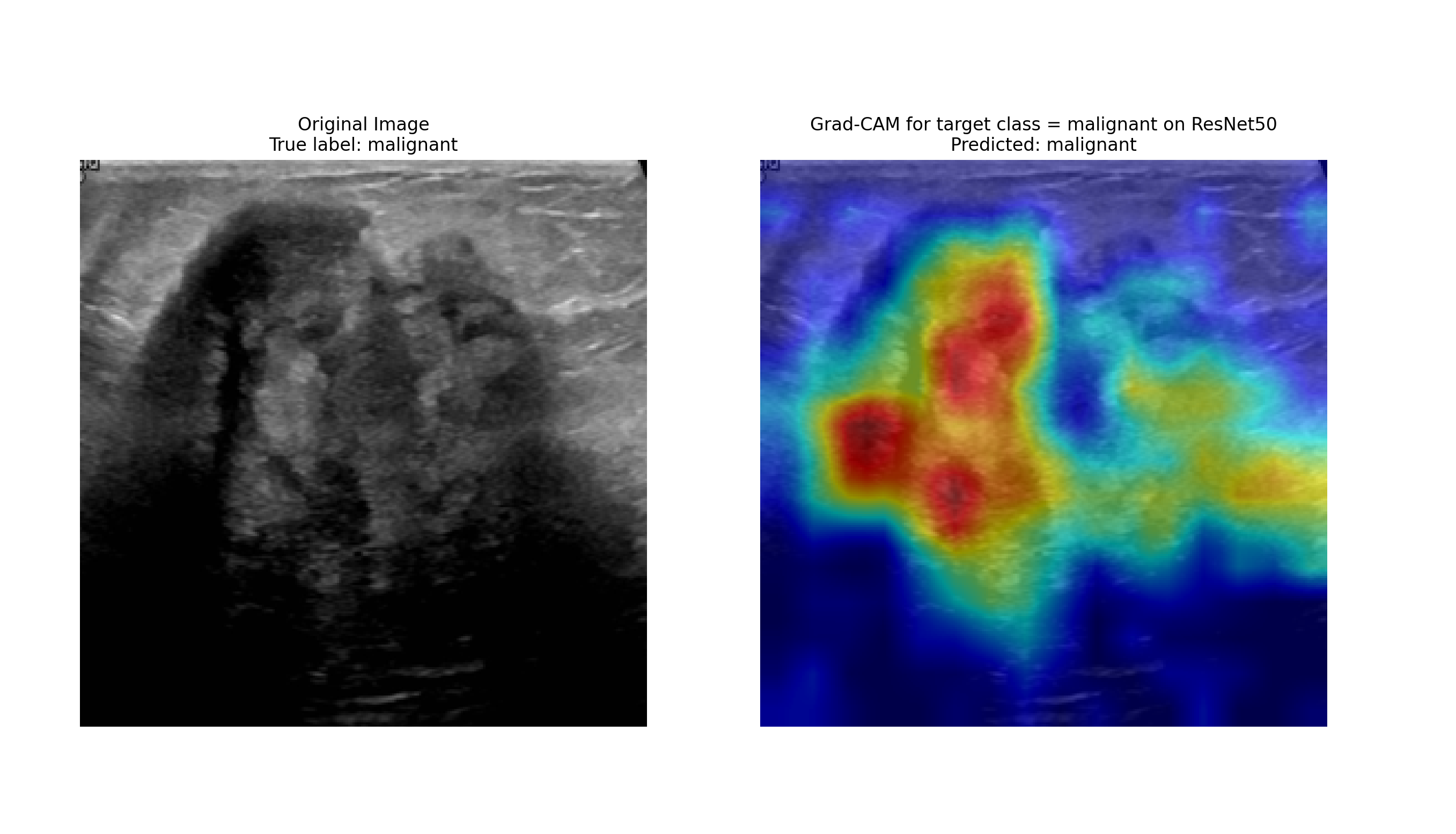}
    \caption{Breast classifier correctly identifies malignancy.}
\end{figure}

\clearpage 

Below is another correct ACL prediction, and another ACL model mistake, visualized with Grad-CAM.

\begin{figure}[!htbp]
    \centering
    \includegraphics[width=0.9\textwidth]{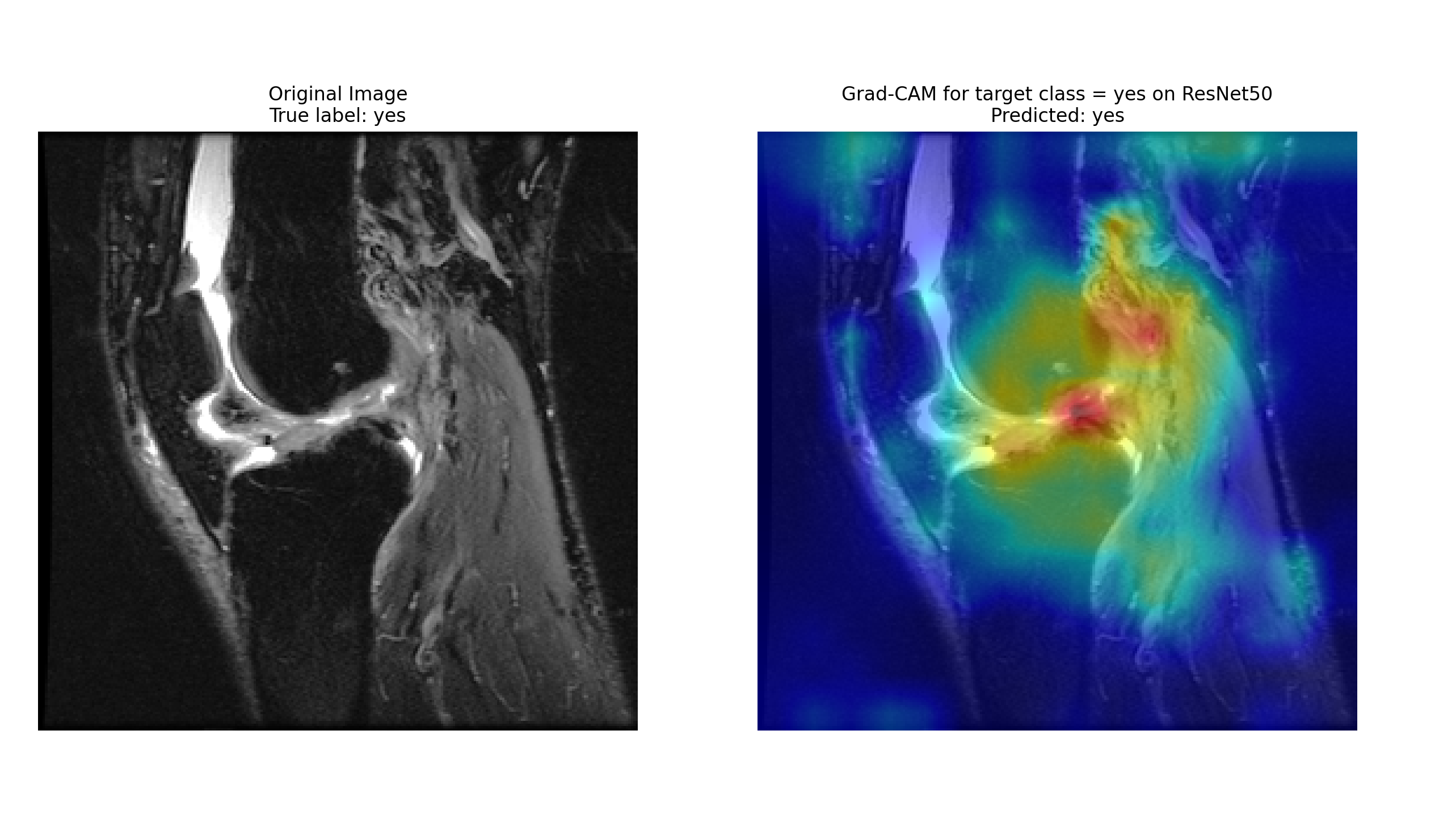}
    \caption{ACL model correctly identifies the torn ACL.}
\end{figure}

\begin{figure}[!htbp]
    \centering
    \includegraphics[width=0.9\textwidth]{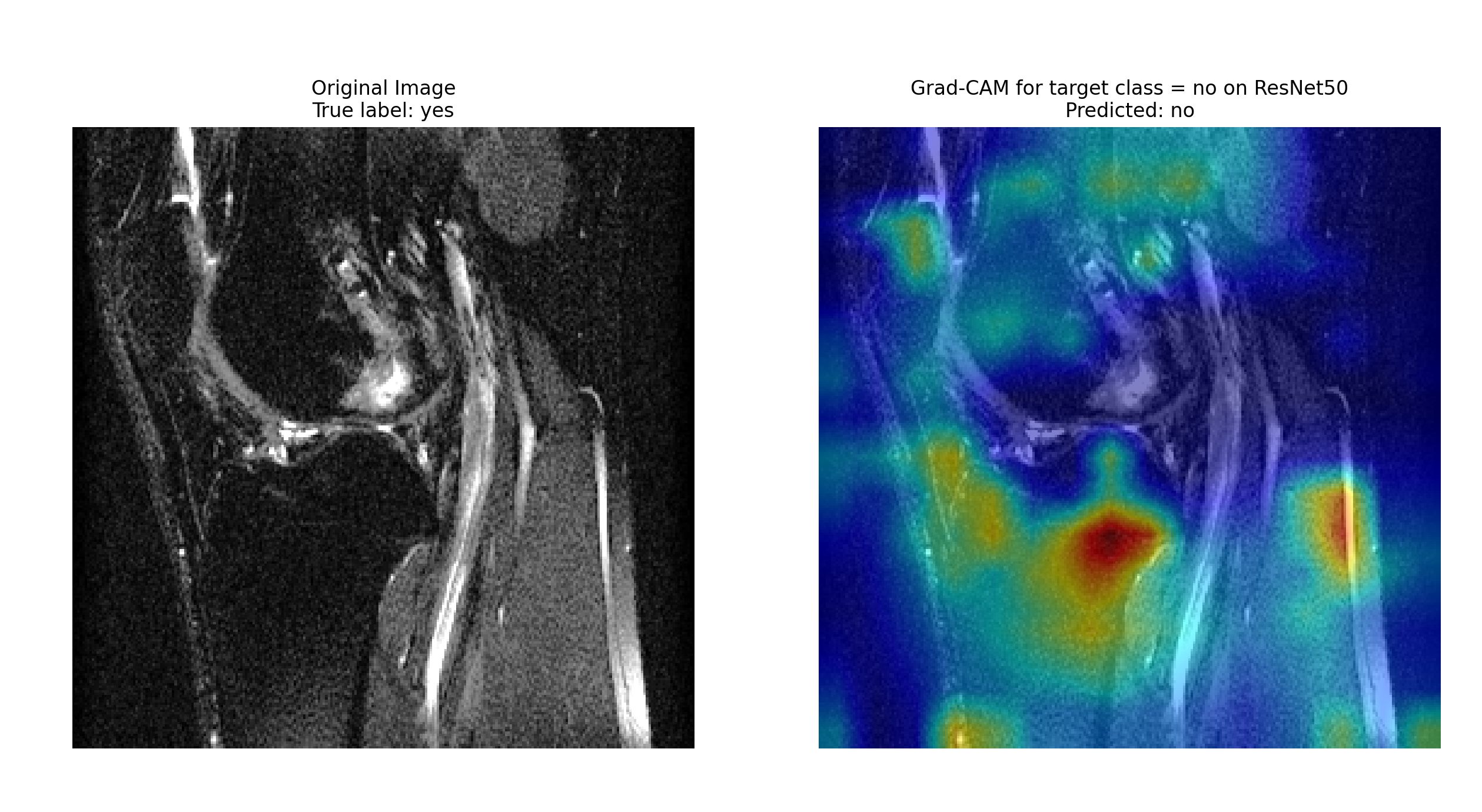}
    \caption{ACL Tear Classification model mistake. The model attention seems to completely break down with this dark, noisy image.}
    \label{fig:acl_gradcam_mistake}
\end{figure}

\clearpage

\section{Baseline Experiments}
\label{sec:baselines}

Our baseline task seeks to replicate published experiments from the RadImageNet GitHub \cite{RadImageNetGitHub}, to ensure initial alignment with their work for equitable comparison. Our experiments compare our updated version of their TensorFlow code (since their code yielded errors when first run), and our PyTorch transfer-learning implementation. 

To establish baselines, we debugged the original TensorFlow implementation from the RadImageNet GitHub repository \cite{RadImageNetGitHub} and developed a corresponding PyTorch implementation. Both frameworks utilized RadImageNet/ImageNet for backbone initialization, and incorporated identical image augmentation and preprocessing techniques alongside linear classifiers. Our goal was to validate the consistency of our implementation with the original by comparing loss and AUC metrics across several TensorFlow and PyTorch experiments, thereby establishing baseline performance for the ACL and Breast tasks without additional architectural enhancements.

\begin{figure}[th]
    \centering
    \includegraphics[width=0.65\linewidth]{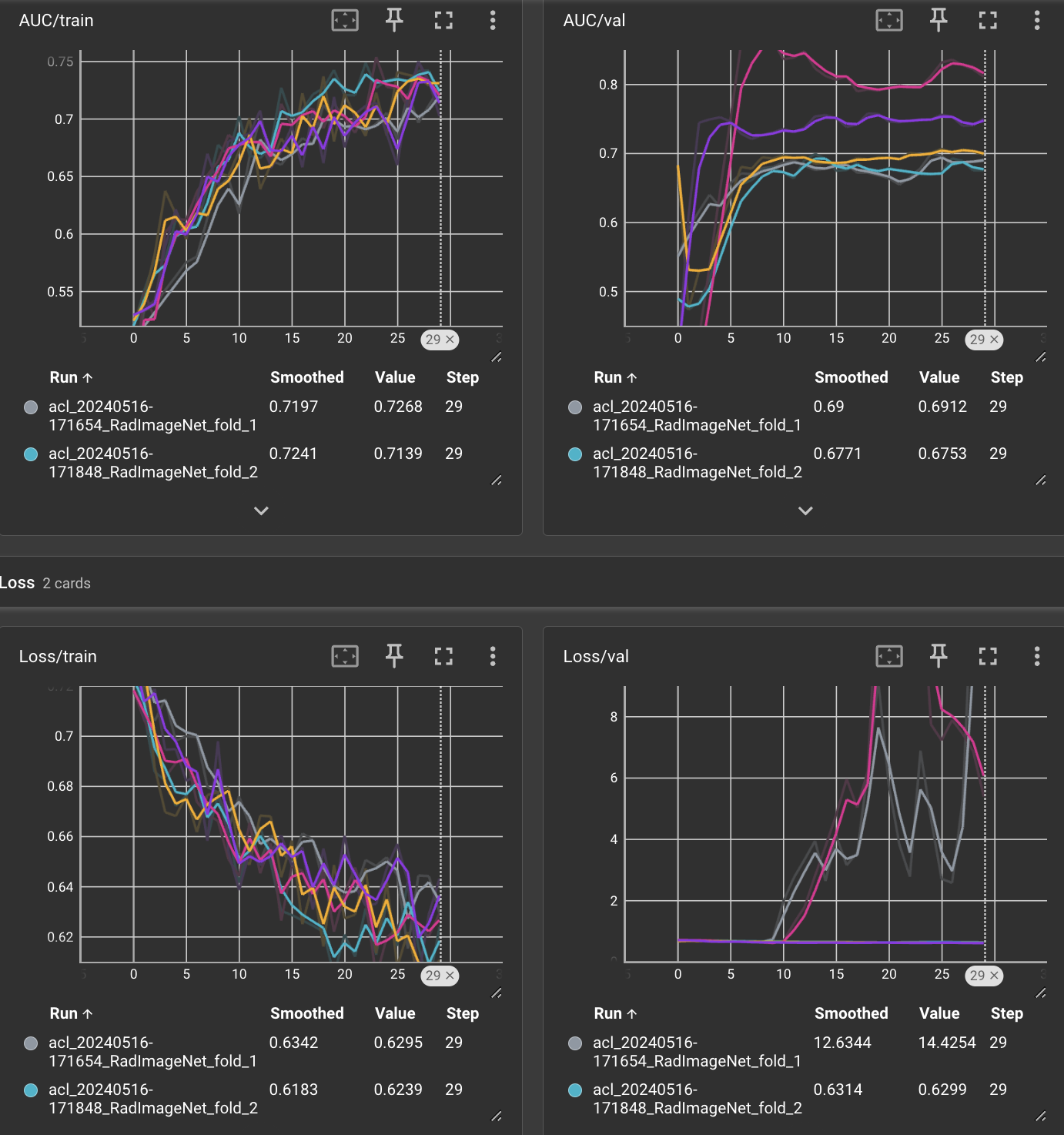}
    \caption{Training instability with RadImageNet weights in InceptionV3.}
    \label{fig:radimage_unstable_inception}
\end{figure}

In \autoref{tab:baseline_results}, we present preliminary results using pretrained RadImageNet weights with ResNet50 and InceptionV3 backbones under various configurations of frozen and unfrozen layers. ResNet50 demonstrated superior convergence and more stable training compared to InceptionV3, which exhibited significant instability when initialized with RadImageNet weights, as illustrated in \autoref{fig:radimage_unstable_inception}. Due to this instability, InceptionV3 was excluded from subsequent experiments.

Training with unfrozen layers significantly enhanced performance; however, the pronounced disparity between training and validation metrics suggests a heightened risk of overfitting, indicating a need for additional regularization strategies. Notably, TensorFlow implementations of InceptionV3 outperformed their PyTorch counterparts, possibly due to subtle differences in TensorFlow's execution and the specific RadImageNet weights used for InceptionV3 in PyTorch. For a detailed analysis of TensorFlow's superior performance, refer to \autoref{subsec:tensorflow_inception_better}.

\clearpage 

\subsection{Loss and AUC across Baseline Experiments}

\begin{table*}[ht]
    \centering
    \caption{Baseline Results}
    \label{tab:baseline_results}
    \begin{tabular}{lcccc}
        \toprule
        & \multicolumn{4}{c}{TensorFlow} \\
        \cmidrule(lr){2-5}
        Experiment & Train Loss & Val Loss & Train AUC & Val AUC \\
        \midrule
        \textbf{ACL ImageNet InceptionV3 Freeze All} & \textbf{0.551433} & 0.538424 & \textbf{0.819705} & 0.819349 \\
        \textbf{ACL RadImageNet InceptionV3 Freeze All} & 0.615819 & 0.607071 & 0.798664 & 0.833382 \\
        \textbf{Breast ImageNet InceptionV3 Freeze All} & 0.555277 &\textbf{ 0.479173} & 0.808335 & \textbf{0.873217} \\
        \textbf{Breast RadImageNet InceptionV3 Freeze All} & 0.571363 & 0.545559 & 0.787512 & 0.829327 \\
        \midrule
        & \multicolumn{4}{c}{PyTorch} \\
        \cmidrule(lr){2-5}
        Experiment & Train Loss & Val Loss & Train AUC & Val AUC \\
        \midrule
        \textbf{ACL ImageNet InceptionV3 Freeze All} & 0.548978 & 0.616359 & 0.796058 & 0.777943 \\
        \textbf{ACL ImageNet ResNet50 Freeze All} & 0.520163 & 0.572959 & 0.831995 & 0.782532 \\
        \textbf{ACL RadImageNet InceptionV3 Freeze All} & 0.646131 & 1.434678 & 0.689376 & 0.746009 \\
        \textbf{ACL RadImageNet ResNet50 Freeze All} & 0.504645 & 0.514543 & 0.854029 & 0.866853 \\
        \textbf{ACL RadImageNet ResNet50 UnfreezeAll} & \textbf{0.007352} &\textbf{ 0.421879} & \textbf{1.000000} & \textbf{0.982571} \\
        \textbf{Breast ImageNet InceptionV3 Freeze All} & 0.495767 & 0.469062 & 0.757553 & 0.814543 \\
        \textbf{Breast ImageNet ResNet50 Freeze All} & 0.443195 & 0.450079 & 0.832770 & 0.847085 \\
        \textbf{Breast RadImageNet InceptionV3 Freeze All} & 0.557808 & 0.515856 & 0.693958 & 0.770700 \\
        \textbf{Breast RadImageNet ResNet50 Freeze All} & 0.508051 & 0.467209 & 0.787071 & 0.831545 \\
        \bottomrule
    \end{tabular}
\end{table*}

\subsection{InceptionV3 training on TensorFlow Performs Better than PyTorch}
\label{subsec:tensorflow_inception_better}

Notably, TensorFlow results are higher than PyTorch results for overlapping architectures. We propose that there are three primary reasons for this: 

\begin{enumerate}
    \item While we exhaustively read through the legacy TensorFlow 2.0 code \cite{tensorflow2015-whitepaper} used by the RadImageNet authors, it was not possible to perfectly match every single detail of the TensorFlow implementation. A large part of the challenge lies in the fact that TensorFlow abstracts much more logic away from the programmer, so we had to do our best to deduce default and hidden behaviors.
    \item The original RadImageNet work primarily focused on TensorFlow implementation, and it is possible that the TensorFlow pre-training weights were similarly given more focus and training time compared with PyTorch. In addition to any architectural differences between the implementations, the experiments are fundamentally different in that TensorFlow relies on RadImageNet's published TF weights, and PyTorch relies on a different set of published PT weights. 
    \item In particular, \texttt{InceptionV3} was highly unstable with the PyTorch RadImageNet weights, which creates a stronger performance gap between the two implementations for RadImageNet pre-training. 
\end{enumerate}

Ideally, as a sanity check of our PyTorch implementation, we would have derived nearly equivalent performance between the two implementations. However, we are confident that we reasonably included all default and hidden behaviors from the legacy TensorFlow 2.0 methods used in the original RadImageNet work into our PyTorch implementation, having traced through the TensorFlow source code in great detail. That said, future investigation into the poor PyTorch performance with InceptionV3 is warranted. 

\clearpage

\section{Additional Image Preprocessing Experiments}
\label{sec:additional_image_prep}
 
 In deriving the aforementioned image preprocessing pipeline, we experimented with several data augmentation techniques on the training data, including random horizontal and vertical flips, random rotation, random affine transformation, and the addition of color jitter and Gaussian blur. These preprocessing steps were also largely informed by the original RadIamgeNet work \cite{Mei2022RadImageNet}. \\
 
Random horizontal and vertical flips, random rotation, and random affine transformation yielded slightly improved model performance on the breast cancer classification and ACL meniscus tear tasks, as measured by validation AUC. However, color jitter and Gaussian blurring decreased performance. We suspect that the improvements in performance from the flips, rotations and affine transformations arises from their artificial inflation / expansion of the training dataset, forcing the model to be more robust. Color jitter (affecting brightness and contrast here, since our images are grayscale) and Gaussian blur both seem to be too destructive for the radiological image signals which determine diagnosis for ACL tears and breast cancer. These transformations might obscure subtle nuances in tissue density or fluid distribution that are key to diagnosis. For example, Gaussian blur reduces the sharpness of important features such as tissue borders, making cancerous growths appear more smooth and oval in shape, which might lead to false 'benign' diagnoses. 
 

 \clearpage

\section{Additional Grid Search Hyperparameter Comparison Box Plots}
\label{subsec:additional_grid_res}

\subsection{Choosing the best Learning Rate Decay, Number of Filters, and Backbone}

Below, we visualize the box + scatter plots used to choose the ideal starter learning rate decay method, number of convolutional filters, and backbone architecture. 

\begin{figure}[!htbp]
    \centering
    \includegraphics[width=0.6\textwidth]{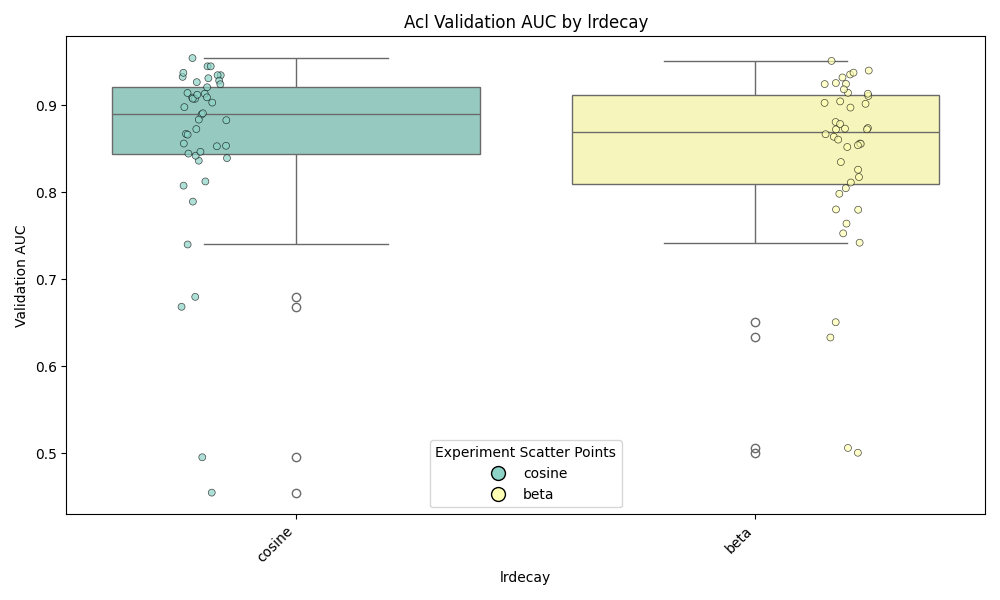}
    \caption{Cosine annealing learning rate decay yields better average validation AUC. }
    \label{fig:overall_val_auc_lrdecay}
    \end{figure}
    
    \begin{figure}[!htbp]
    \centering
    \includegraphics[width=0.6\textwidth]{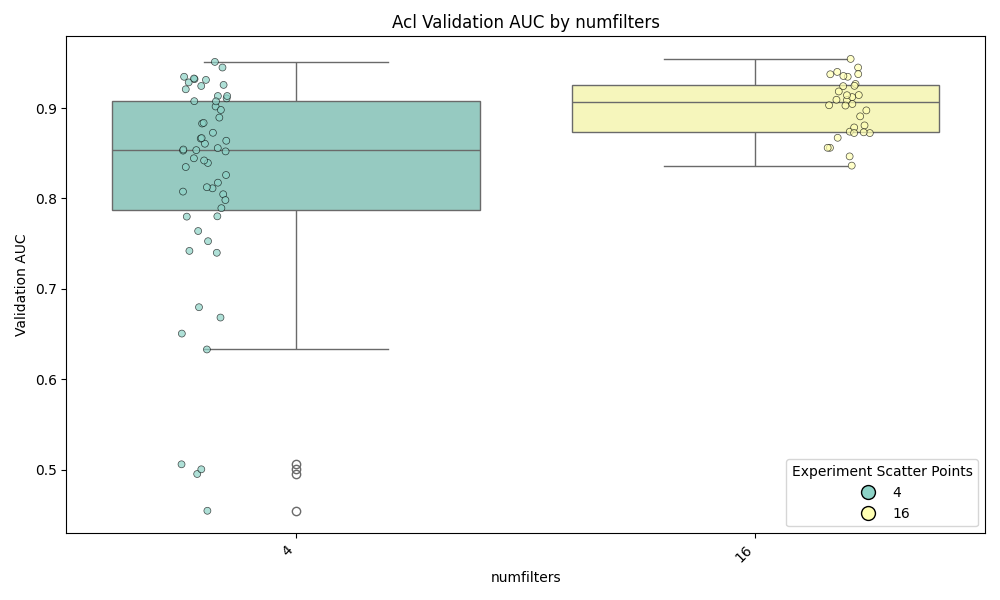}
    \caption{16 convolutional filters yields better average performance overall.}
    \label{fig:overall_val_auc_numfilters}
    \end{figure}
    
    \begin{figure}[!htbp]
    \centering
    \includegraphics[width=0.6\textwidth]{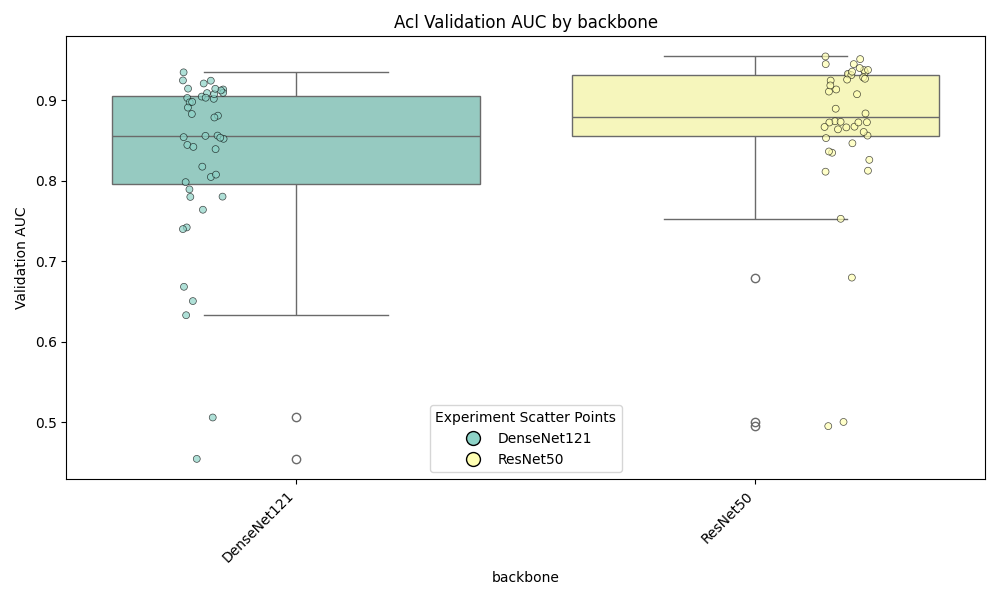}
    \caption{Validation AUC distribution is superior overall for ResNet 50.}
    \label{fig:overall_val_auc_backbone}
    \end{figure}

\subsection{Grid Search Results Hold for Multiple Metrics}

Below, we observe that the optimal hyperparameters and architectures we suggest in this work are consistently ranked across not just validation AUC, but also rank nearly the same across validation F1 and accuracy. The exception is that the pure \texttt{Conv} classifier now wins out on average (but not in peak performance) compared to \texttt{ConvSkip}.

\begin{figure}[!htbp]
    \centering
    \includegraphics[width=0.6\textwidth]{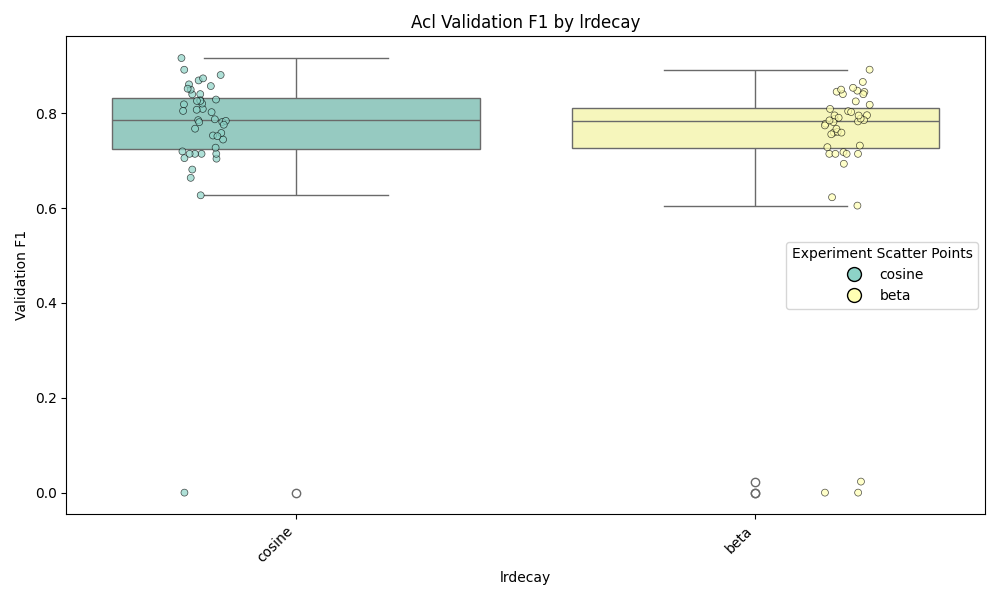}
    \caption{Cosine annealing learning rate decay yields better average validation F1 score.}
    \label{fig:overall_val_f1_lrdecay}
\end{figure}

\begin{figure}[!htbp]
    \centering
    \includegraphics[width=0.6\textwidth]{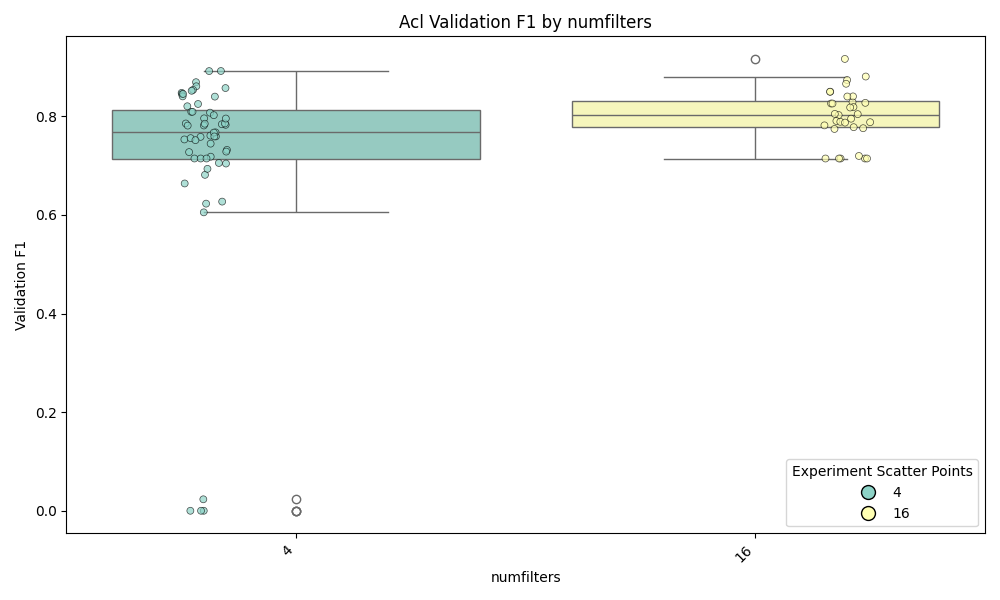}
    \caption{16 convolutional filters yield better average performance overall.}
    \label{fig:overall_val_f1_numfilters}
\end{figure}

\begin{figure}[!htbp]
    \centering
    \includegraphics[width=0.6\textwidth]{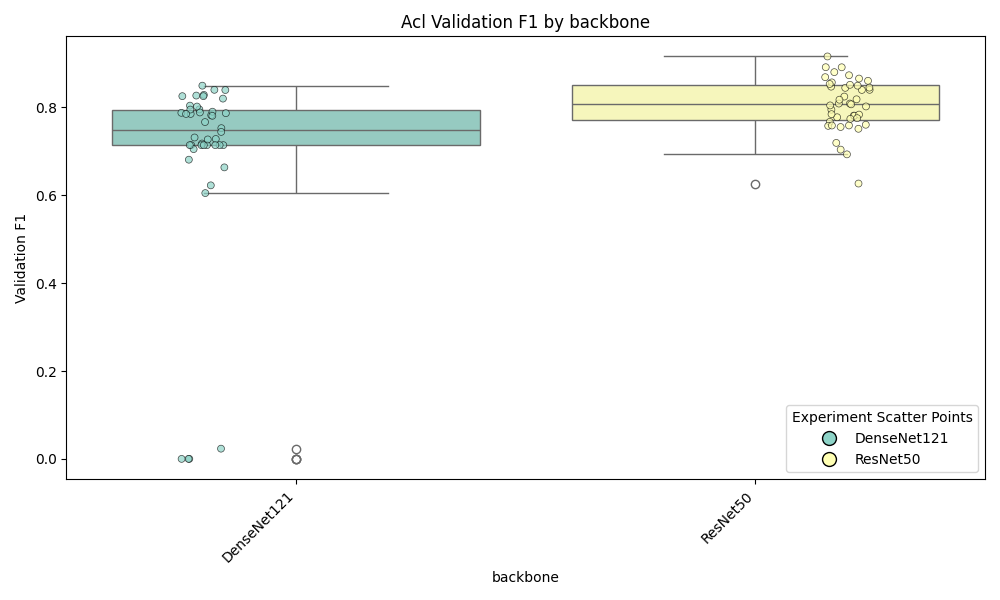}
    \caption{Validation F1 score distribution is superior overall for ResNet 50.}
    \label{fig:overall_val_f1_backbone}
\end{figure}

\begin{figure}[htbp]
    \centering
    \includegraphics[width=0.6\textwidth]{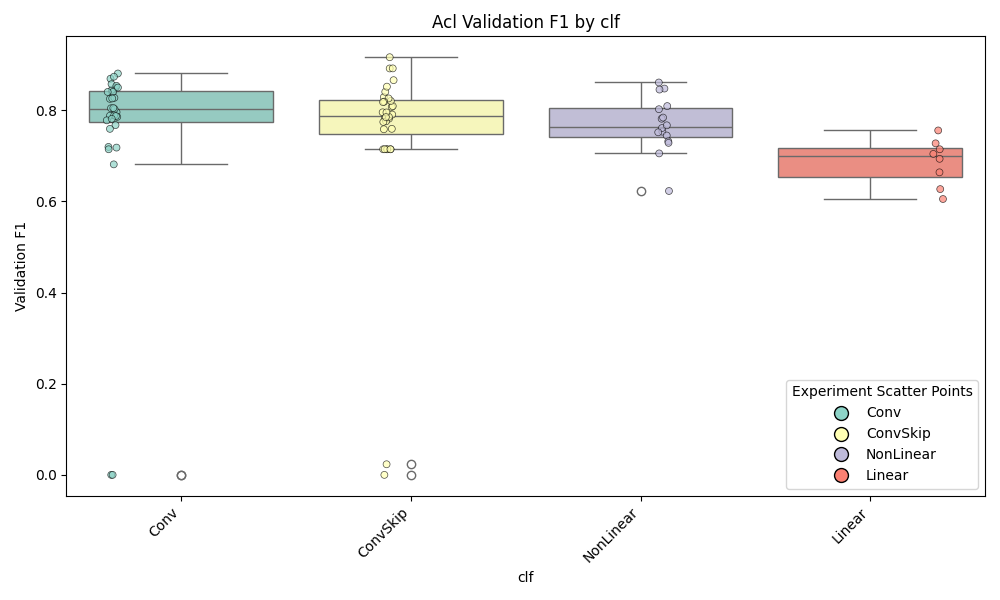}
    \caption{Conv yields better average validation F1, but ConvSkip yieldes best peak validation F1.}
    \label{fig:overall_val_f1_clf}
\end{figure}

\begin{figure}[!htbp]
    \centering
    \includegraphics[width=0.6\textwidth]{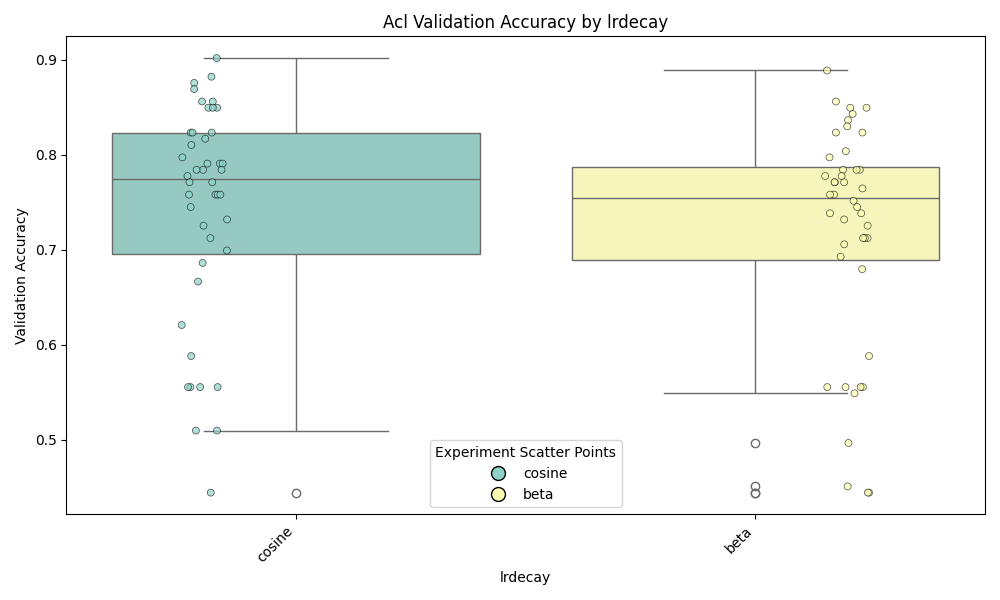}
    \caption{Cosine annealing learning rate decay yields better average validation Accuracy.}
    \label{fig:overall_val_accuracy_lrdecay}
\end{figure}

\begin{figure}[!htbp]
    \centering
    \includegraphics[width=0.6\textwidth]{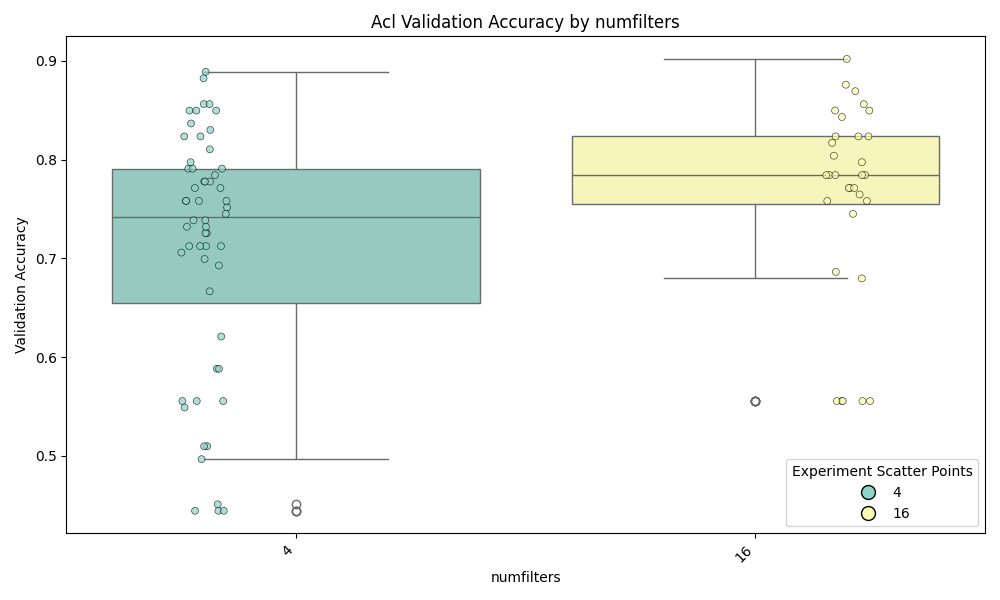}
    \caption{16 convolutional filters yield better average performance overall.}
    \label{fig:overall_val_accuracy_numfilters}
\end{figure}

\begin{figure}[!htbp]
    \centering
    \includegraphics[width=0.6\textwidth]{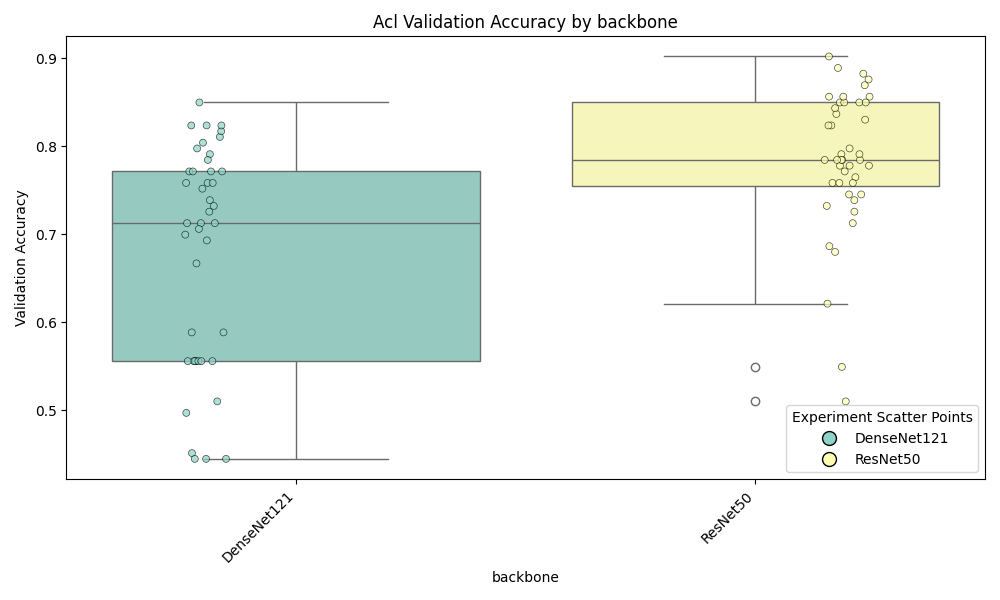}
    \caption{Validation Accuracy distribution is superior overall for ResNet 50.}
    \label{fig:overall_val_accuracy_backbone}
\end{figure}

\begin{figure}[htbp]
    \centering
    \includegraphics[width=0.6\textwidth]{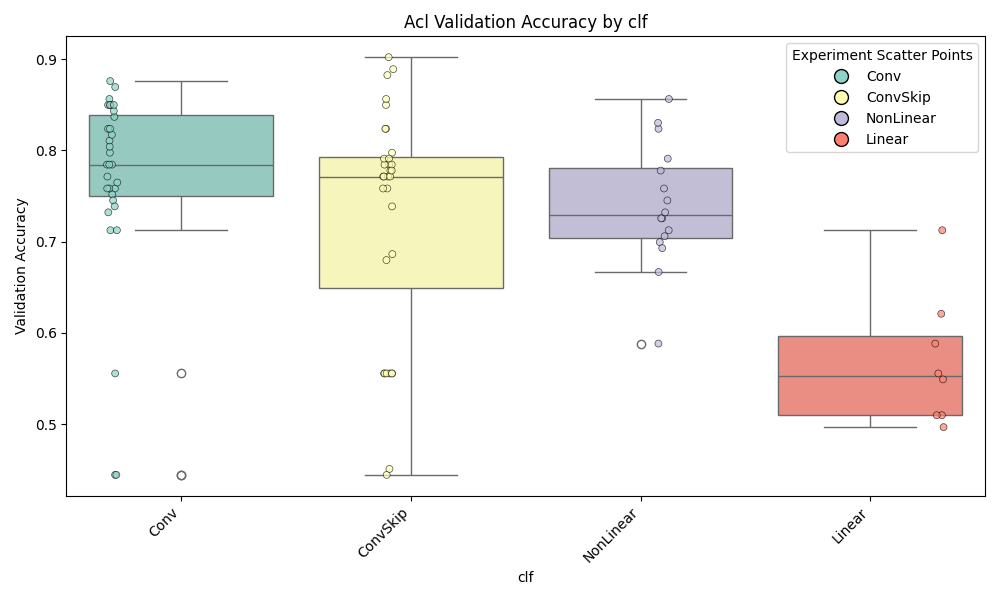}
    \caption{Conv yields better average validation accuracy, but ConvSkip yieldes best peak validation accuracy.}
    \label{fig:overall_val_accuracy_clf}
\end{figure}

\clearpage

\subsection{Full Linear Mixed Model Results}

\begin{table}[!htbp] \centering 
  \caption{Full Linear Mixed Model for Pretraining} 
  \label{tab:full_lmm_pretraining}
\begin{tabular}{@{\extracolsep{5pt}}lc} 
\\[-1.8ex]\hline 
\hline \\[-1.8ex] 
 & \multicolumn{1}{c}{\textit{Results}} \\ 
\cline{2-2} 
\\[-1.8ex] & Fixed Effects \\ 
\hline \\[-1.8ex] 
 Intercept & Estimate: 0.738, SE: 0.088, df: 4.58, t: 8.341, p: 0.0006*** \\ 
 PretrainRadImageNet & Estimate: 0.016, SE: 0.008, df: 167, t: 1.938, p: 0.0543. \\ 
\hline \\[-1.8ex] 
 & Random Effects \\ 
\hline \\[-1.8ex] 
 clf & Variance: 0.02370, Std.Dev.: 0.15394 \\ 
 numfilters & Variance: 0.0002393, Std.Dev.: 0.01547 \\ 
 lrdecay & Variance: 0.00006962, Std.Dev.: 0.008344 \\ 
 backbone & Variance: 0.001712, Std.Dev.: 0.04137 \\ 
 task & Variance: 0.001690, Std.Dev.: 0.04111 \\ 
 Residual & Variance: 0.003158, Std.Dev.: 0.05620 \\ 
\hline \\[-1.8ex] 
Observations & \multicolumn{1}{c}{176} \\ 
REML Criterion & \multicolumn{1}{c}{-471.7} \\ 
Scaled Residuals & \multicolumn{1}{c}{Min: -3.4535, 1Q: -0.6725, Median: 0.0553, 3Q: 0.6873, Max: 4.5190} \\ 
\hline 
\hline \\[-1.8ex] 
\textit{Note:}  & \multicolumn{1}{r}{$^{***}$p$<$0.001; $^{**}$p$<$0.01; $^{*}$p$<$0.05; $^{.}$p$<$0.1} \\ 
\end{tabular} 
\end{table}

\begin{table}[!htbp] \centering 
  \caption{Full Linear Mixed Model for Classifier} 
  \label{tab:full_lmm_clf}
\begin{tabular}{@{\extracolsep{5pt}}lc} 
\\[-1.8ex]\hline 
\hline \\[-1.8ex] 
 & \multicolumn{1}{c}{\textit{Results}} \\ 
\cline{2-2} 
\\[-1.8ex] & Fixed Effects \\ 
\hline \\[-1.8ex] 
 Intercept & Estimate: 0.829, SE: 0.044, df: 2.48, t: 18.777, p: 0.000983*** \\ 
 clfConvSkip & Estimate: 0.018, SE: 0.010, df: 167, t: 1.788, p: 0.0755. \\ 
 clfLinear & Estimate: -0.313, SE: 0.016, df: 165, t: -19.135, p: <2e-16*** \\
 clfNonLinear & Estimate: -0.038, SE: 0.013, df: 157, t: -2.954, p: 0.0036** \\
\hline \\[-1.8ex] 
 & Random Effects \\ 
\hline \\[-1.8ex] 
 task & Variance: 0.00169, Std.Dev.: 0.04111 \\ 
 pretrain & Variance: 0.00009892, Std.Dev.: 0.009946 \\
 backbone & Variance: 0.001712, Std.Dev.: 0.04137 \\ 
 lrdecay & Variance: 0.00006961, Std.Dev.: 0.008343 \\
 numfilters & Variance: 0.0002303, Std.Dev.: 0.015175 \\
 Residual & Variance: 0.003159, Std.Dev.: 0.05620 \\
\hline \\[-1.8ex] 
Observations & \multicolumn{1}{c}{176} \\ 
REML Criterion & \multicolumn{1}{c}{-475.7} \\ 
Scaled Residuals & \multicolumn{1}{c}{Min: -3.4584, 1Q: -0.6522, Median: 0.0902, 3Q: 0.7033, Max: 4.5135} \\ 
\hline 
\hline \\[-1.8ex] 
\textit{Note:}  & \multicolumn{1}{r}{$^{***}$p$<$0.001; $^{**}$p$<$0.01; $^{*}$p$<$0.05; $^{.}$p$<$0.1} \\ 
\end{tabular} 
\end{table}
    
\clearpage

\subsection{Wilcoxon Test Results}
\label{subsec:wilcoxon}

The inherent positive correlation within our dataset's hyperparameter groups biases tests towards rejecting the null hypothesis. This predisposition means that while outcomes of the test must be interpreted with a lot of care, a failure to reject the null can still provide some information. Consequently, to evaluate if the performance of models pretrained on ImageNet significantly differs from those pretrained on RadImageNet, we utilized the Wilcoxon Rank Sum test which is robust to non-normally distributed data. The results from this test (see\autoref{tab:combined_test_results}) show that the differences in AUC for the breast cancer detection task are not statistically significant (\textit{p} = 0.8023), suggesting no substantial advantage of either pretraining method in this context. In contrast, the significant result for the ACL tear detection task (\textit{p} = 0.003935) suggests a potential superiority of RadImageNet, which warrants further investigation. This test's bias toward rejecting the null hypothesis amplifies the importance of these non-significant findings, particularly for the breast cancer task where no advantage is detected. These outcomes therefore inform our understanding of pretraining efficacy across tasks within the limited framework of a brief, 5-epoch training regimen.

    \begin{table}[h]
    \centering
    \caption{Wilcoxon Rank Sum Test Results for ImageNet vs. RadImageNet Pretraining}
    \label{tab:combined_test_results}
    \begin{tabular}{@{}lcc@{}}
    \toprule
    \textbf{Metric} & \textbf{Breast Task} & \textbf{ACL Task} \\ \midrule
    W-statistic & 998.5 & 622 \\
    p-value & 0.8023 & 0.003935** \\
    \textit{Note:}  & \multicolumn{2}{r}{$^{**}$p$<$0.01} \\ \bottomrule
    \end{tabular}
    \end{table}
\clearpage

\subsection{McNemar Tests}
\label{subsec:mcnemar}
McNemar's Chi-squared tests (see \autoref{tab:mcnemar_results}) reveal a significant difference in predictions only for the ACL task, likely influenced by the smaller sample size of breast task's test set. These findings confirm that the advantages of ImageNet pretraining are more pronounced in ACL model performance compared to the breast task.

\begin{table}[htbp]
\centering
\caption{McNemar's Chi-squared Test Results}
\label{tab:mcnemar_results}
\begin{tabular}{@{\extracolsep{5pt}}lcc}
\toprule
& \textbf{ACL Task} & \textbf{Breast Task} \\ 
\midrule
\textbf{McNemar's $\chi ^2$} & 10.562 & 0.9 \\
\textbf{Degrees of Freedom} & 1 & 1 \\
\textbf{p-value} & 0.001154 & 0.3428 \\
\bottomrule
\end{tabular}
\end{table}

\clearpage

\section{Hardware}
\label{sec:hardware}

All final experiments were performed using the Metal Performance Shaders backend on an Apple Silicon M3 Max GPU with 1 TB of storage, 30-core GPU (300GB/s memory bandwidth), and 36GB of unified memory.

Experimentation on Google Compute Engine with $1 \times$Tesla T4 GPU and cuda PyTorch optimizations proved to be much slower, even after implementing automatic mixed-precision training. AMP sped up GCP performance by about $2x$, but training was still a few times slower than MPS due to significantly fewer GPU cores and less VRAM. 
\clearpage

\section{Analysis of Original RadImageNet Work}
\label{sec:critique}

The primary objective of the RadImageNet paper \cite{Mei2022RadImageNet} was to evaluate the differential performance, as measured by AUC values, between models pre-trained on RadImageNet and those pre-trained on ImageNet. The RadImageNet authors use DeLong test \cite{DeLong1988} for comparing AUC metrics across different models, as we do here. However, the observation that the original paper's 95\% confidence intervals sometimes include zero difference between RadImageNet and ImageNet pre-training, despite extremely low p-values (i.e., $<0.001$), suggests a potential methodological discrepancy in how these metrics were derived——possibly indicating that confidence intervals were computed using bootstrapping methods \cite{Efron1979}. However, the author's state that both the p-value and the confidence intervals come from the DeLong test, which seemingly violates core statistical principles, namely the duality of confidence intervals and $t$-tests.

This discrepancy does not inherently undermine the validity of the results but does raise critical questions regarding the statistical robustness of the comparisons and the conclusions drawn. Firstly, the absence of clarity on the comparative methodology for AUC values—-whether the tests applied were suitable for this dataset-—leaves a significant gap in the reproducibility and interpretation of the findings. Secondly, the method by which the number of AUC values was determined remains unclear. If these values are the outcome of a grid search across hyperparameter settings, it is essential to ascertain whether the same optimal hyperparameter space applies equally to models pre-trained on ImageNet and RadImageNet. The selection of specific hyperparameter subspace for comparison may introduce biases or misrepresentations. As we note in the present work, our conclusion that ImageNet exhibits superior performance may merely be an artifact of the architectural and hyperparameter search space we investigate. This could equally be true of the original RadImageNet work. Alternatively, if the AUC values are derived from consistent fine-tuning settings across varying training and validation datasets, the appropriateness of the chosen fine-tuning hyperparameters for each model—particularly whether those for the ImageNet model were as judiciously selected as those for the RadImageNet model—remains an open question.

To establish more robust conclusions, further research should explicitly clarify the hypothesis testing methods used, ensure hyperparameter optimization is consistent across models, and validate the selection of statistical methods for computing confidence intervals and p-values.

\clearpage

\section{Additional Best Model Information}
\subsection{Additional Learning Curve Visualizations}

In training best models for the ACL and breast tasks, we noted that ImageNet yielded a better upper bound on performance in our experimentation, as measured by validation AUC. Notably, the learning curves also demonstrate superior convergence across other useful metrics, such as F1 and accuracy. 

\begin{figure}[!htbp]
    \centering
    \includegraphics[width=0.5\textwidth]{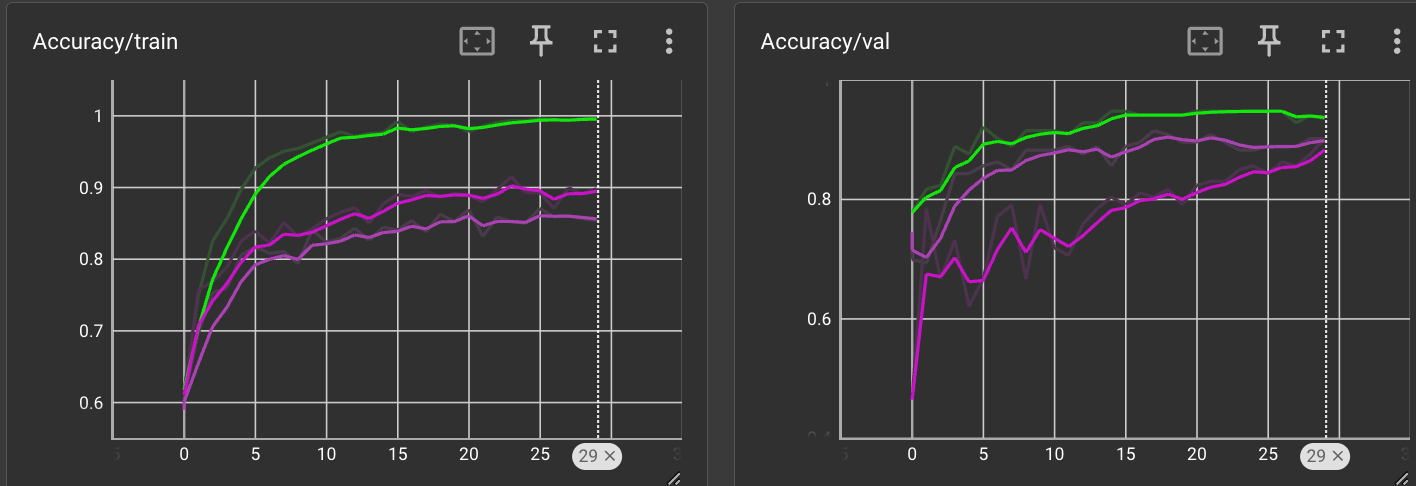}
    \includegraphics[width=0.5\textwidth]{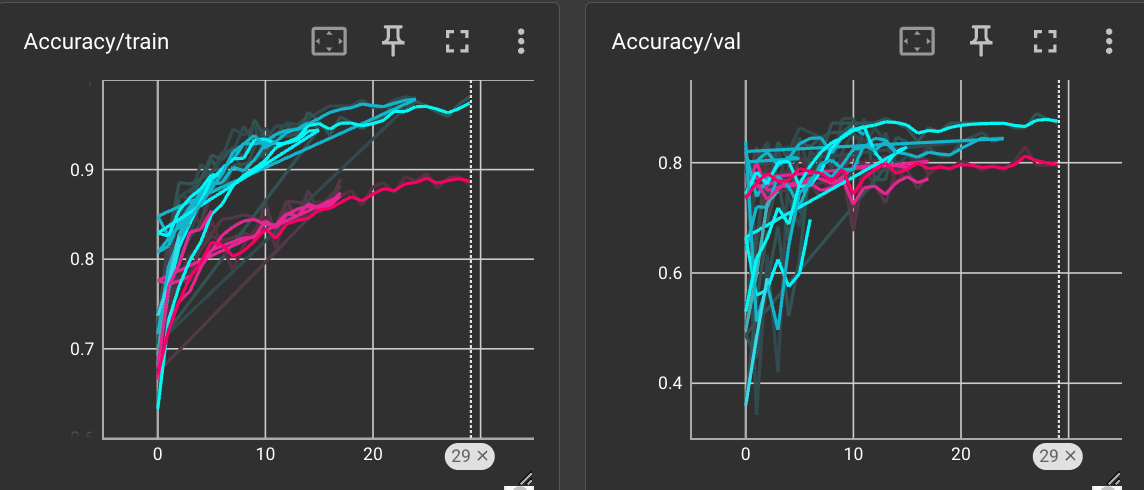}
    \caption{Validation Accuracy caps out for RadImageNet initialized best models (pink), whereas ImageNet initialized ACL models (green) and ImageNet initialized breast models (blue) achieve better convergence.}
    \label{fig:radimagenet_caps_out_acc}
\end{figure}

\begin{figure}[!htbp]
    \centering
    \includegraphics[width=0.5\textwidth]{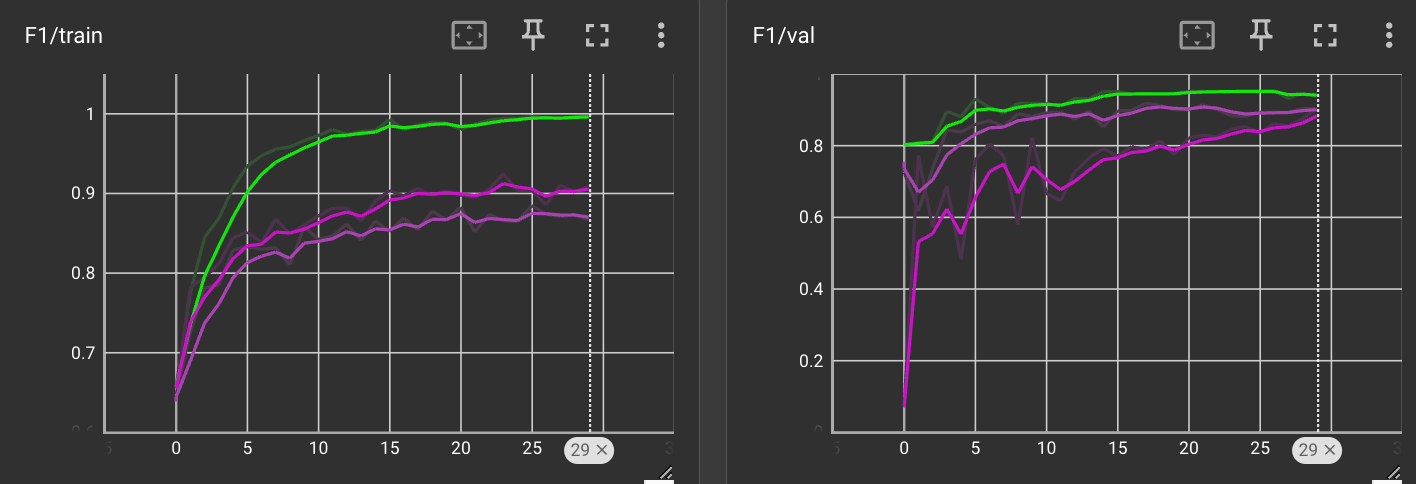}
    \includegraphics[width=0.5\textwidth]{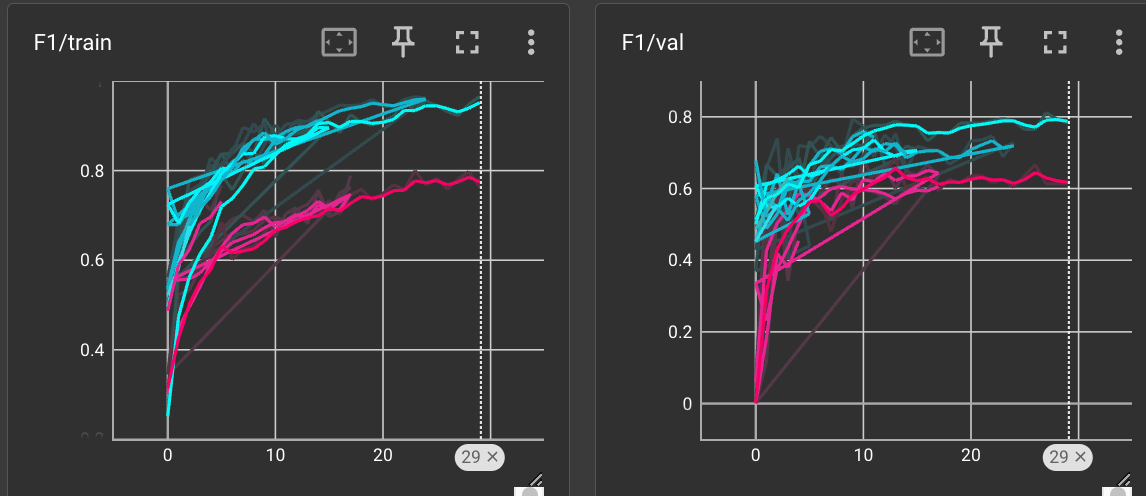}
    \caption{Validation F1 score caps out for RadImageNet initialized best models (pink), whereas ImageNet initialized ACL models (green) and ImageNet initialized breast models (blue) achieve better convergence.}
    \label{fig:radimagenet_caps_out_f1}
\end{figure}
\clearpage

\end{document}